\pdfoutput=1

\documentclass[11pt]{article}

\usepackage{acl}

\usepackage{times}
\usepackage{latexsym}

\usepackage{kotex}
\usepackage[para]{threeparttable}
\usepackage{booktabs}
\usepackage{multirow}
\usepackage{xspace}
\usepackage{graphicx}

\newcommand{\teal}[1]{\textcolor{teal}{#1}}
\newcommand{\purple}[1]{\textcolor{purple}{#1}}

\newcommand{\up}[1]{\teal{\textbf{#1}}}
\newcommand{\down}[1]{\purple{\underline{#1}}}

\newcommand{\ours}{HyperCLOVA\xspace}

\usepackage[T1]{fontenc}

\usepackage[utf8]{inputenc}

\usepackage{microtype}

\title{On the Effect of Pretraining Corpora on\\ In-context Learning by a Large-scale Language Model}

\author{Seongjin Shin\thanks{\ \ Equal contribution.}\ $^{,1}$\ \ \  Sang-Woo Lee$^{*,1,2}$\ \ \  Hwijeen Ahn$^{1}$\ \ \  Sungdong Kim$^{2}$\\
\textbf{HyoungSeok Kim$^{1}$\ \ \ Boseop Kim$^{1}$\ \ \ Kyunghyun Cho$^{3}$\ \ \ Gichang Lee$^{1}$}\\
\textbf{Woomyoung Park$^{1}$\ \ \ Jung-Woo Ha$^{1,2}$\ \ \ Nako Sung$^{1}$\ \ \ }\\
\\
NAVER CLOVA$^{1}$\ \ \  NAVER AI Lab$^{2}$\ \ \ NYU$^{3}$}

\begin{document}
\maketitle

\begin{abstract}
Many recent studies on large-scale language models have reported successful in-context zero- and few-shot learning ability.
However, the in-depth analysis of when in-context learning occurs is still lacking.
For example, it is unknown how in-context learning performance changes as the training corpus varies.
Here, we investigate the effects of the source and size of the pretraining corpus on in-context learning in \ours, a Korean-centric GPT-3 model. 
From our in-depth investigation, we introduce the following observations: (1) in-context learning performance heavily depends on the corpus domain source, and the size of the pretraining corpus does not necessarily determine the emergence of in-context learning, 
(2) in-context learning ability can emerge when a language model is trained on a combination of multiple corpora, even when each corpus does not result in in-context learning on its own,
(3) pretraining with a corpus related to a downstream task does not always guarantee the competitive in-context learning performance of the downstream task, especially in the few-shot setting,
and (4) the relationship between language modeling (measured in perplexity) and in-context learning does not always correlate: e.g., low perplexity does not always imply high in-context few-shot learning performance.
\end{abstract}

\section{Introduction}
NLP community has been surprised by emergence of in-context learning ability of a large-scale language model (LM) such as GPT-3~\cite{brown2020language} despite no duplication between downstream task data and the pretraining corpus. Indeed, in-context learning uses a natural language description and a few examples to prime a language model. Then the language model can predict the answer of a new example without updating the language model's parameters.
Since the release of GPT-3, various large-scale in-context language models have been proposed \cite{gpt-neo,kim2021changes,zeng2021pangu,rae2021scaling,hoffmann2022training,chowdhery2022palm}.

There still remain many questions on language models' in-context learning capability despite these successful reports.
For example, the relationship between the choice of a pretraining corpus and downstream in-context learning task accuracy is unknown.
Previous studies argue pretraining with the corpus similar to the downstream task improves the downstream performance, but these observations are often limited to the case
where a pretrained language model is finetuned for the downstream task
\cite{gururangan2020don,lee2020biobert,micheli2020importance}.

In addition, analysis on the relation between the validation perplexity of a language model and in-context learning performance is still less investigated.
Previous research on in-context learning implicitly assumes that perplexity is predictive of in-context learning performance by showing scaling law property of their model \cite{kaplan2020scaling,brown2020language,kim2021changes}. \citet{rae2021scaling} also use perplexity for the hyperparameter selection on corpus reweighting in the pretraining of their in-context learner. However, their explicit correlations are less discovered.

Motivated by this lack of in-depth analysis on the relationship between in-context learning and corpus properties,
we vary the sources and sizes of pretraining corpora and analyze their impact on in-context learning, using \ours, which is a Korean-centric large LM.~\cite{kim2021changes}. 
We mainly discover in-context few-shot learning as in the previous work \cite{kim2021changes} but also explore in-context zero-shot learning.
We use HyperCLOVA corpus, which is a large-scale pretraining corpus mainly in Korean collected by \citet{kim2021changes}, as a base corpus from which we derive pretraining corpora for our experiments.

Our major findings include:
\begin{itemize}
    \item \textbf{Corpus Source:} In-context learning performance depends heavily on corpus sources, and with some sources, in-context learning does not work effectively.
    For example, the model trained only on a subcorpus of blog (\texttt{Blog}) achieves competitive in-context few-shot learning performance, but training on a subcorpus of community website (\texttt{Cafe}) or online news articles (\texttt{News}) hardly yields in-context few-shot learning ability.
    \item \textbf{Corpus Combination:} In-context learning ability can emerge by fusing two corpora, even when each on its own does not result in in-context learning.
    For example, while training only on \texttt{KiN} corpus, which consists of QnA websites, or \texttt{Ency} corpus, which consists of Encyclopedia websites, in-context few-shot learning ability was not observed, but training on both corpora makes the emergence of in-context few-shot learning.
    \item \textbf{Domain Relevance:} 
    Pretraining with a corpus related to a downstream task seems to help in-context zero-shot learning performance, but is not indicative of the competitive in-context few-shot learning performance.
    For example, training on only \texttt{News} corpus makes a relatively good in-context zero-shot learning ability on a news-related downstream task, e.g., news topic classification based on its title, KLUE-YNAT \cite{park2021klue}, but does not yield in-context few-shot learning ability.
    \item \textbf{Perplexity:} 
    Although perplexity and in-context learning accuracies correlate well when training a single model, perplexity alone does not reflect the difference in in-context learning accuracies across different language models. This is prominent particularly when they were trained using different pretraining corpora.
    For example, \texttt{Cafe} model, the model trained with \texttt{Cafe} corpus, has the second lowest validation perplexity on various domain sources after \texttt{Blog} model, but fails to emerge in-context few-shot learning.
\end{itemize}

\section{Related Work}

\subsection{In-context Learning}

\citet{brown2020language} demonstrate the concept of in-context learning, where a few training examples and/or task descriptions are provided together with a new input for a large-scale LM to produce a target of this input, without requiring any parameter update. A few training examples are used in the in-context few-shot learning setting, whereas no training example is used in the in-context zero-shot setting. A few follow-up studies have tried to improve the in-context learning ability~\cite{Zhao2021CalibrateBU, holtzman-etal-2021-surface}. 
On the other hand, another group of papers tries to explain the mechanism of in-context few-shot learning \cite{min2022rethinking,xie2022an}.

\subsection{Domain Relevance on Pretraining Corpus}
Previous studies argue a better downstream accuracy is observed with a pretraining corpus more similar to the downstream task corpus \cite{gururangan2020don,lee2020biobert,micheli2020importance}.
However, these observations are limited to the case where a pretrained language model is finetuned for the downstream task. 

There are a few studies on the effects of different corpus on the relationship between pretraining and in-context learning.
A notable example is Codex, where GPT-3 is trained on Github corpus so that the model can generate code from comments \cite{chen2021evaluating}. 
However, the corpus used for Codex is limited to code comments and the corresponding code.
We study the effect of pretraining corpus on in-context learning performance using various domains.

\subsection{Quantity and Quality of Pretraining Corpus}
There have been several studies on the quantity and quality of pretraining data. 
\citet{raffel2020exploring} conduct an ablation study on different pretraining corpus on T5, and their filtered C4 corpus makes T5 perform better in downstream tasks. 
As with GPT-3, researchers generally improve the quality of their language model through data filtering \cite{brown2020language,kim2021changes}.  
Our research differs from the existing work in that we focus on in-depth analysis of how the amount of data and the corpus source affect in-context learning.

\subsection{Multi-task Learning}
Multi-task learning approaches, which explicitly finetune on the in-context learning objective by using numerous NLP tasks, are proposed recently to tackle zero/few-shot transfer to the unseen task at test time~\cite{wei2021finetuned, sanh2021multitask, chen2021metalearning, min2021metaicl}.

Unlike the studies in a finetuning paradigm, many properties of the in-context learning related to pretraining corpus are still unknown.
As the previous multi-task studies show that diverse tasks improve the ability of in-context learning, our study shows that diverse pretraining corpora strengthen the ability of in-context learning.

\section{Task Definition}

\subsection{Model}
\label{subsec:model}
We use the variants of \ours with various parameter sizes and pretraining corpus. We mainly experiment with models with 1.3B parameters, but we also include the result for 6.9B-sized models. 
All models have a maximum sequence length of 2,048.

We emphasize that all models use the same vocabulary across all our experiments.
We use the morpheme-aware byte-level BPE tokenizer trained with HyperCLOVA corpus \cite{kim2021changes} for all models. We train multiple models with different portions of HyperCLOVA corpus to investigate the effects of the source and size of the corpus on in-context learning ability.

\subsection{Pretraining with Different Corpus}
\label{sec:pretraining_corpus}

\begin{table}[t!]
    \centering
    \small
    \setlength\tabcolsep{4.5pt}
    \begin{threeparttable}
    \begin{tabular}{llr}
        \toprule
        Name& Description & Tokens\\
        \midrule
        \texttt{Blog} & Blog corpus & 273.6B \\
        \texttt{Cafe} & Online community corpus & 83.3B \\
        \texttt{News} & News corpus & 73.8B \\
        \texttt{Comments} & Crawled comments corpus & 41.1B \\
        \texttt{KiN} & Korean QnA website corpus & 27.3B\\
        \texttt{Modu} & Collection of five datasets  & 6.0B \\
        \texttt{Ency} & Encyclopedia corpus & 1.7B\\
        \texttt{Others} & Other corpus & 55.0B\\
        \midrule
        Total & & 561.8B \\
        \bottomrule
    \end{tabular}
    \end{threeparttable}
    \caption{Descriptions of HyperCLOVA corpus \cite{kim2021changes}.}
    \label{table:corpus}
\end{table}

We analyze the effect of seven subcorpora in the HyperCLOVA corpus: \texttt{Blog}, \texttt{Cafe}, \texttt{News}, \texttt{Comments}, \texttt{KiN}, \texttt{Modu}, and \texttt{Ency}. Table \ref{table:corpus} summarizes the characteristics of the subcorpora.
\texttt{Blog}, \texttt{Cafe}, and \texttt{News} are taken from blog, community sites, and online news articles of NAVER\footnote{https://www.naver.com/}, a Korean web portal service, respectively. \texttt{Comments} is the comment threads related to the three subcorpora mentioned above. 
\texttt{KiN} comes from NAVER's online community QnA service similar to Quora. \texttt{Ency} is a collection of encyclopedic texts including Korean Wikipedia. \texttt{Modu} consists of five public datasets constructed by  National Institute of the Korean Language\footnote{https://corpus.korean.go.kr/}, including 3.2B of news, 2.1B of written language, 0.4B of spoken language, 0.2B of web corpus, and 0.02B tokens of messenger.
\texttt{Others} was excluded to investigate the explicit effects of domain corpus sources on in-context learning because \texttt{Others} is the corpus where various subcorpora are taken from multiple heterogeneous sources.
Tables \ref{table:corpus-example} and \ref{table:corpus-example-en} in Appendix show the examples of seven pretraining corpus in Korean and English, respectively.
\texttt{ALL} denotes the original HyperCLOVA corpus including \texttt{Others}.

For corpora with less than 150B tokens, we assign 99\% of each corpus to the pretraining corpus and randomly extract 10,000 examples from the remaining 1\% to the validation corpus for measuring validation perplexity. For corpora with more than 150B tokens, we make the training corpora 150B tokens via random sampling and construct a validation set with 10,000 examples randomly sampled from the remaining. 
As a result, the maximum training set size of each corpus is 150B tokens.

The validation set for each corpus consists of 10,000 examples and is used for the early stopping of models trained with each corpus.
However, we combine all validation set to make the entire validation set contains 70,000 examples for seven domains, and the entire validation set is used for calculating perplexity, as described in Section \ref{subsec:measuring-validation-perplexity}.

\subsection{Downstream Tasks}
We evaluate in-context learning performance of each corpus-specific model on four Korean downstream task datasets used in \citet{kim2021changes}: NSMC\footnote{https://github.com/e9t/nsmc}, KorQuAD \cite{lim2019korquad1}, AI Hub translation\footnote{https://aihub.or.kr/aidata/87}, and YNAT \cite{park2021klue}. 
NSMC is a binary sentiment classification dataset on movie review. KorQuAD is a machine reading comprehension dataset similar to SQuAD 1.0 \cite{rajpurkar-etal-2016-squad}. AI Hub translation dataset consists of Korean-English parallel sentences from news, government websites, legal documents, etc. YNAT is a topic classification problem with seven classes.

\begin{figure}[t] 
\centering
\includegraphics[width=0.99\columnwidth]{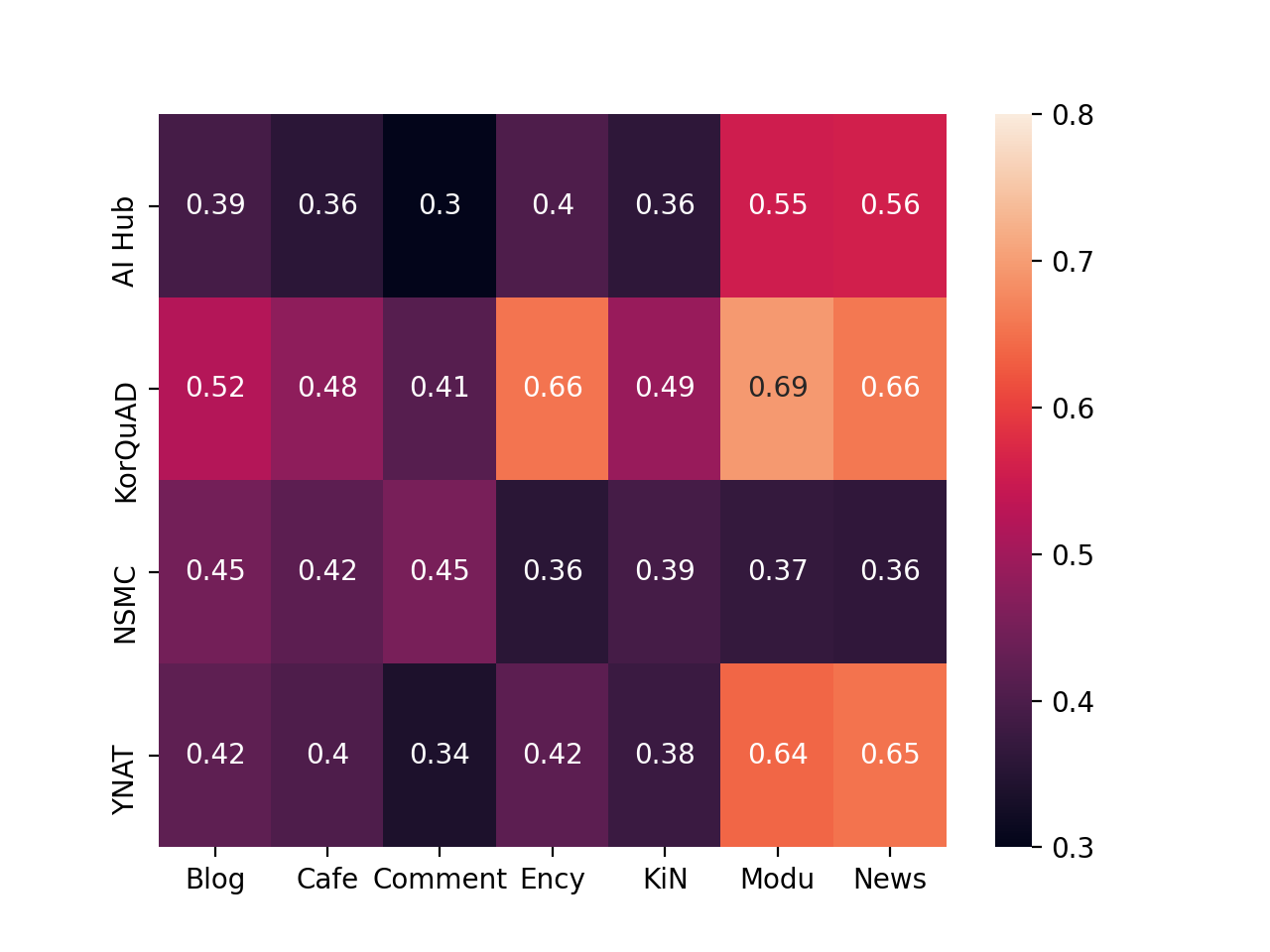}
\caption{Vocabulary overlap ratio between pretraining corpus and downstream task. Top 1,000 nouns are used to calculate the ratio. Nouns are extracted using our in-house part-of-speech tagger.}
\label{fig:vocab-overlap}
\end{figure}

We think that three datasets for downstream tasks are closely related to the HyperCLOVA corpus.
Passages which construct KorQuAD are taken from Korean Wikipedia, which is also a part of the \texttt{Ency}. 
YNAT is a topic classification task of news headlines, so the downstream task is deeply related to the \texttt{News} corpus. 
A significant portion of parallel sentences for AI Hub translation dataset also comes from news articles.
\texttt{KiN} corpus is also related to the translation task. About 2.5\% of QnA data in \texttt{KiN} includes Korean questions on the English language, as a foreign language. These question-style passages often include Korean-English sentence pairs in the passage.
Vocabulary overlap between downstream tasks and HyperCLOVA corpus is depicted in Figure~\ref{fig:vocab-overlap}.

\subsection{Experimental Details}
We try our best to make the same hyperparameter of \citet{kim2021changes}, including global batch size, training step, maximum sequence length, learning rate, and so on.
In our experiments, the models are trained for 72K steps with a global batch size of 1,024. We note that under this setting, the number of tokens that were actually used in pretraining is 150B. Therefore, we set the maximum size of training corpus to 150B as in Section \ref{sec:pretraining_corpus}.

In most experiments, validation perplexity decreases monotonically as training goes on. Thus, we use the checkpoint at 72K step. 
The only exception is the \texttt{Ency} model. The \texttt{Ency} model has a minimum validation loss at 12K steps, which is likely to be caused by overfitting to pretraining data due to a small size of the data. Therefore, we use early-stopping checkpoints at the 12K steps for the report.

For optimization, AdamW \cite{loshchilov2018decoupled}  with the learning rate of 2.0e-4 and the cosine learning rate scheduling are used. We use the mixed precision training. Models are trained on the Nvidia Superpod which consists of 1,024 A100 GPUs spread across 128 nodes.
Using Superpod, it spends around 18 hours to train 1.3B model with 72K steps.

For classification tasks such as NSMC and YNAT, we use a rank classification approach \cite{wei2021finetuned}, where we compare pre-defined outputs (``positive'' and ``negative'') and take the one with higher probability. KorQuAD and AI Hub are free-form completion tasks, where we directly generate output tokens using the greedy decoding.

In the few-shot experiments, the number of shots is set to 70, 4, 4, and 70 for NSMC, KorQuAD, AI Hub, and YNAT, respectively. 
Downstream tasks are performed 12, 1, 3, and 6 times with different random seeds for NSMC, KorQuAD, AI Hub, and YNAT, respectively. We report the average performance. Random seed influences the sampling of shots from training data and their order. 
The reason KorQuAD has only one random seed is described in Appendix \ref{sec:appendix-examples-of-few-shot-prompt}.
Appendix \ref{sec:appendix-examples-of-few-shot-prompt} also includes the examples of the few-shot prompts used in our experiments.
These all experimental settings in the few-shot experiments, from the number of shots to the number of random trials, basically come from the experimental setting of \citet{kim2021changes}. However, we change the number of trials of YNAT from 3 to 6, because we found that the standard derivation of YNAT is relatively high.

\subsection{Measuring Validation Perplexity}
\label{subsec:measuring-validation-perplexity}

We report validation perplexity in various tables and figures to verify our argument. We use the term ``PPL'' to denote validation perplexities on the validation set. The validation set consists of 70,000 examples from seven corpus sources, as described in Section \ref{sec:pretraining_corpus}. We emphasize that, for calculating PPL, all experiments use \textbf{the same vocabulary and validation set}.

In Tables \ref{table:result1} and \ref{table:result2}, we use \textit{Italic} font for the results from a multi-domain model, which is pretrained with two or more mixed corpora. Since a multi-domain model trains more domains than a single-domain model, the PPLs of multi-domain models are generally lower than those of single-domain models. To keep readers from directly comparing PPLs between a single-domain and a multi-domain model, we use italic font for the results of a multi-domain model.

\begin{table*}[t!]
    \centering
    \small
    \begin{threeparttable}
    \begin{tabular}{lcccccccc}
        \toprule
        \multirow{2}{*}{Model} & Corpus & \multirow{2}{*}{PPL} & NSMC & \multicolumn{2}{c}{KorQuAD} & \multicolumn{2}{c}{AI Hub (BLEU)} & YNAT \\
        & Train  & & (Acc)  & (EM) & (F1) & Ko$\rightarrow$En & En$\rightarrow$Ko & (F1) \\
        \midrule
        \midrule
        Majority & - & -  & 50.35 & 0.0 & 0.0 & 0.0 & 0.0 & 8.26\\
        \midrule
        \texttt{ALL} & 150B  & \textit{119.99} & 84.59 & 56.17 & 73.47 & 6.15 & 23.36 & 59.57  \\
        \texttt{ALL} w/o \texttt{Others} & 150B &\textit{119.66} & 84.59 & 56.49 & 74.20 & 6.14 & 23.21 & 50.76 \\
        \midrule
        \texttt{Blog} & 150B & 152.40 & \up{83.50} & \up{50.74} & \up{69.34} & \up{3.82} & \up{20.11} & \up{60.68}  \\
        \texttt{Cafe} & 82.5B  & 170.85 & \down{57.77} & \down{3.12} & \down{14.26} & \down{2.83} & \up{16.53} & \down{11.04} \\
        \texttt{News} & 73.1B  & 234.78 & \down{50.72} & \down{0.14} & \down{9.96} & \down{1.10} & \up{15.88} & \down{14.36} \\
        \texttt{Comments} & 40.7B & 225.39 & \up{79.78} & \down{14.69} & \down{33.33} & \down{0.79} & \down{5.06} & \up{36.17} \\
        \texttt{KiN} & 27.0B & 187.80 & \down{54.73} & \down{4.85} & \down{18.99} & \up{6.81} & \up{18.16} & \down{9.23} \\ 
        \texttt{Modu} & 5.9B & 226.01 & \up{69.91} & \up{30.20} & \up{49.29} & \down{1.21} & \down{6.13} & \up{43.27} \\
        \texttt{Ency} & 1.7B & 549.40 & \down{53.81} & \down{0.71} & \down{11.88} & \down{0.58} & \down{0.69} & \down{27.99} \\
        \midrule
        \texttt{Blog 54B} & 54.0B & 155.69 & \up{83.06} & \up{49.13} & \up{68.10} & \up{3.93} & \up{21.12} & \up{57.97} \\
        \midrule
        \texttt{Blog 27B} & 27.0B & 165.60 & \up{80.27} & \down{10.91} & \down{23.41} & \up{5.35} & \up{12.32} & \up{48.19} \\
        \texttt{Cafe 27B} & 27.0B & 169.81 & \down{49.91} & \down{1.37} & \down{13.98} & \up{4.25} & \up{20.74} & \down{8.60} \\
        \texttt{News 27B} & 27.0B & 239.79 & \down{50.64} & \down{0.80} & \down{8.02} & \down{2.42} & \up{15.78} & \down{27.20} \\
        \texttt{Comments 27B} & 27.0B & 229.65 & \up{80.50} & \down{13.02} & \down{31.53} & \down{1.70} & \down{3.28} & \down{25.79} \\
        \bottomrule
    \end{tabular}
    \end{threeparttable}
    \caption{In-context few-shot learning performance with different pretraining corpus. Models with 1.3B parameters are used. 
    Majority means classifying each label with the primary class, and its score is 0 for KorQuAD and AI Hub. \down{Purple-underline} denotes the score is below the mean performance value of \texttt{ALL} and Majority baseline, and \up{Teal-bold} denotes the score is above.}
    \label{table:result1}
\end{table*}

\begin{table*}[t!]
    \centering
    \small
    \begin{threeparttable}
    \begin{tabular}{lcccccccc}
        \toprule
        \multirow{2}{*}{Model} & Corpus & \multirow{2}{*}{PPL} & NSMC & \multicolumn{2}{c}{KorQuAD} & \multicolumn{2}{c}{AI Hub (BLEU)} & YNAT \\
        & Train  & & (Acc)  & (EM) & (F1) & Ko$\rightarrow$En & En$\rightarrow$Ko & (F1) \\
        \midrule
        \midrule
        Majority & - & -  & 50.35 & 0.0 & 0.0 & 0.0 & 0.0 & 8.26\\
        \midrule
        \texttt{ALL} & 150B  & \textit{119.99} & \down{61.68} & \up{56.17} & \up{73.47} & \up{7.43} & \up{24.81} & \up{42.79} \\
        \midrule
        \texttt{Blog} & 150B & 152.40 & \up{75.28} & \up{50.74} & \up{69.34} & \up{5.44} & \up{22.88} & \up{49.34}  \\
        \texttt{Cafe} & 82.5B  & 170.85 & \up{69.38} & \down{3.12} & \down{14.26} & \up{4.34} & \up{16.44} & \up{38.12} \\
        \texttt{News} & 73.1B  & 234.78 & \down{54.96} & \down{0.14} & \down{9.96} & \down{1.28} & \down{10.21} & \up{48.03} \\
        \texttt{Comments} & 40.7B & 225.39 & \down{57.69} & \down{14.69} & \down{33.33} & \down{1.98} & \down{3.94} & \down{32.48} \\
        \texttt{KiN} & 27.0B & 187.80 & \down{65.43} & \down{4.85} & \down{18.99} & \up{4.64} & \down{10.42} & \up{36.06} \\ 
        \texttt{Modu} & 5.9B & 226.01 & \up{72.50} & \up{30.22} & \up{49.30} & \down{2.39} & \down{7.55} & \up{35.28} \\
        \texttt{Ency} & 1.7B & 549.40 & \down{42.96} & \down{14.01} & \down{31.51} & \down{0.80} & \down{0.77} & \down{30.22} \\
        \midrule
    \end{tabular}
    \end{threeparttable}
    \caption{In-context zero-shot performance with different pretraining corpus.}
     \label{table:result1-zero}
\end{table*}

\section{Experimental Results}

We perform intensive experiments to answer these four main questions:
\begin{enumerate}
    \item How large do the source and the size of pretraining corpora have the effects on emerging in-context learning ability? (Section \ref{subsec:result-source} and \ref{subsec:result-size})
    \item What is the effect of combining various corpora? (Section \ref{subsec:result-combining})
    \item How large does domain relevance of corpus influence on model performances of the downstream task? (Section \ref{subsec:result-relevance}) 
    \item How strong is the correlation between validation perplexity and in-context learning of language models? (Section \ref{subsec:result-perplexity})
\end{enumerate}

\subsection{Main Results}
Tables \ref{table:result1} and \ref{table:result2} show the in-context few-shot results on various pretraining corpus sources and different corpus combination, respectively. Tables \ref{table:result1-zero} and \ref{table:result2-zero} depict the in-context zero-shot results of some models in Tables \ref{table:result1} and \ref{table:result2}, respectively.
All results in Tables \ref{table:result1}, \ref{table:result1-zero}, \ref{table:result2}, and \ref{table:result2-zero} come from models with 1.3B parameters.
Tables \ref{table:result1-std} and \ref{table:result2-std} in Appendix \ref{sec:appendix-details-experimental-results} present the standard derivation values on the results of Tables \ref{table:result1} and \ref{table:result2}.

In Tables \ref{table:result1}, \ref{table:result1-zero}, \ref{table:result2}, and \ref{table:result2-zero}, \down{Purple-underline} denotes the score is below the mean performance value of \texttt{ALL} and Majority baseline in Table \ref{table:result1}, and \up{Teal-bold} denotes the score is above. We use this mean value of Majority and \texttt{ALL} in Table \ref{table:result1} as the performance basis to prevent the in-context learning performance of each model from being distorted by the high basis performance of two classification tasks such as NSMC and YNAT.

Tables \ref{table:result1} and \ref{table:result3} include in-context few-shot results on various pretraining corpus sizes. 
In Table \ref{table:result3}, for example, 56B and 6B correspond to the 1/10 and 1/100 of the original HyperCLOVA corpus with 560B tokens, respectively. The 56B tokens and 6B tokens models are trained with around 3 and 25 epochs, respectively, so that both models can be trained with 72K training steps.
On the other hand, Table \ref{table:result1} compares 27B tokens models trained with different corpus sources to show the results in controlled corpus size.

\subsection{Effect of Corpus Source}
\label{subsec:result-source}

It is noticeable that in-context learning ability emerges differently depending on pretraining corpus sources, as shown in Tables \ref{table:result1}, \ref{table:result1-zero}, \ref{table:result2}, and \ref{table:result2-zero}.
For example, \texttt{Blog} model makes competitive in-context few-shot learning performance to \texttt{ALL} model, while each of \texttt{Cafe} and \texttt{News} models hardly shows in-context few-shot learning ability from Table \ref{table:result1}. 
It is also noticeable that \texttt{Modu} model performs better than \texttt{Cafe} and \texttt{News} model although the size of \texttt{Modu} corpus is less than 1/10 of \texttt{Cafe} or \texttt{News} corpus, showing the corpus size is not the only factor to predict in-context learning performance.
Likewise, it is also interesting that \texttt{Cafe}+\texttt{News} model also shows poor performance despite the same size to \texttt{Blog} and \texttt{ALL}, as shown in Table~\ref{table:result2}.

These differences in in-context learning are dramatic compared to the finetuning results we expect in general. For a comparative experiment between in-context learning and finetuning in our setting, we also finetuned the experimented models with LoRA \cite{hu2021lora}. As Table \ref{table:lora} in Appendix \ref{sec:lora} shows, the performance differences in finetuning are much smaller than in the case of in-context learning.

\subsection{Effect of Corpus Size}
\label{subsec:result-size}

Table \ref{table:result3} shows that reducing the corpus size from 150B to 56B does not decrease the performance severely despite training with 1/10 of corpus.
However, the performance degradation of 6B tokens model is remarkable comparing to \texttt{ALL} model. Nevertheless, it is noticeable that 6B tokens model still performs much better than \texttt{Cafe}+\texttt{News} model, which trains 150B tokens of \texttt{Cafe} and \texttt{News} corpus.

We can also see the similar results for three Blog models of different sizes in Table \ref{table:result1}. 
\texttt{Blog} and \texttt{Blog} 54B achieve similar performance. However, like in \texttt{ALL} 6B, \texttt{Blog} 27B performs quite worse than \texttt{Blog} 54B.

\begin{figure}[t]
    \centering
    \includegraphics[width=0.49\linewidth]{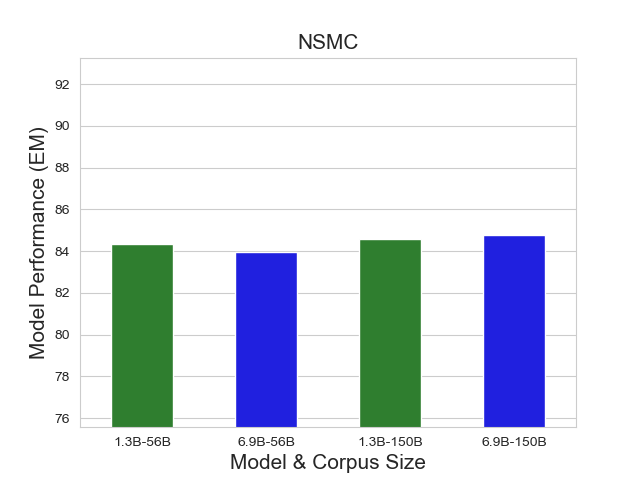}
    \includegraphics[width=0.49\linewidth]{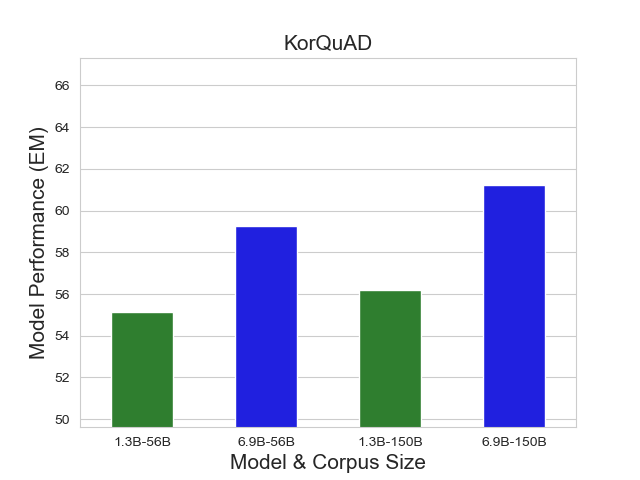}
    \includegraphics[width=0.49\linewidth]{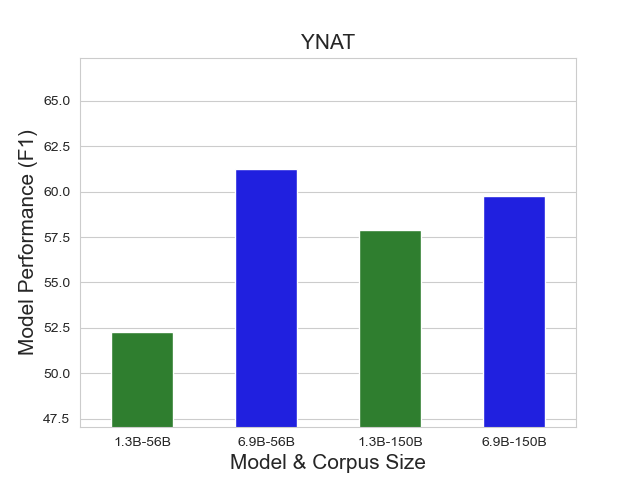}
    \caption{Comparison on two model sizes and two corpus sizes of the original HyperCLOVA corpus such as 1.3B-sized model and 6.9B-sized model, and 56B and 150B tokens. Few-shot results are reported. Green rectangle denotes 1.3B-sized model and blue rectangle denotes 6.9B-sized model.}
    \label{fig:model_size_vs_downstream_task}
\end{figure}

Figure \ref{fig:model_size_vs_downstream_task} shows the comparison between 1.3B-sized model and 6.9B-sized model. 
In the 6.9B-sized models, the in-context few-shot performance with 56B tokens does not decrease significantly compared to 150B tokens, as in the 1.3B-sized models.

\subsection{Effect of Combining Corpora}
\label{subsec:result-combining}
One of our main goals is to investigate the effects of combining multiple corpora from various sources on in-context learning performance. 
Table \ref{table:result2}  shows that in-context few-shot learning ability can be emerged by combining two corpora, even if each of both corpora cannot provide in-context few-shot learning ability. 
For example, \texttt{KiN}+\texttt{Ency} model succeeds to make in-context learning ability in most tasks, while each of \texttt{KiN} and \texttt{Ency} fails in most tasks. 
Likewise, \texttt{Cafe}+\texttt{KiN} model succeeds to make in-context few-shot learning ability, while each of \texttt{Cafe} and \texttt{KiN} fails in most tasks. 
In-context zero-shot abilities of these models follow similar patterns as shown in Table \ref{table:result2-zero}.

\begin{table*}[t!]
    \centering
    \small
    \begin{threeparttable}
    \begin{tabular}{lcccccccc}
        \toprule
        \multirow{2}{*}{Corpus Type} & Corpus & \multirow{2}{*}{PPL} & NSMC & \multicolumn{2}{c}{KorQuAD} & \multicolumn{2}{c}{AI Hub (BLEU)} & YNAT \\
        & Train  & & (Acc)  & (EM) & (F1) & Ko$\rightarrow$En & En$\rightarrow$Ko & (F1) \\
        \midrule
        \midrule
        \texttt{ALL} & 150B & \textit{119.99} & 84.59 & 56.17 & 73.47 & 6.15 & 23.36 & 59.57  \\
        \midrule
        \midrule
        \multicolumn{8}{l}{\textbf{The Case where In-context few-shot learning Emerges by Combining Two Poor Corpora}} \\
        \midrule
        \texttt{KiN}+\texttt{Ency} & 28.7B & \textit{164.69} &  \down{59.17} & \up{42.09} & \up{61.00} & \up{8.99} & \up{23.12} & \up{42.84}  \\
        \texttt{Cafe}+\texttt{KiN} & 109.5B & \textit{141.92} & \up{76.42} & \up{38.45} & \up{59.00} & \up{8.41} & \up{23.41} & \up{56.96}\\
        \midrule
        \midrule
        \multicolumn{8}{l}{\textbf{The Case where In-context few-shot learning Does Not Emerge by Combining Two Poor Corpora}} \\
        \midrule
        \texttt{Cafe}+\texttt{News} & 150B & \textit{154.20} & \down{54.15} & \down{8.95} & \down{22.72} & \up{4.45} & \up{17.77} & \down{8.19} \\
        \midrule
        \midrule
        \multicolumn{8}{l}{\textbf{The Case of Combining In-context few-shot Emerging Corpora}}\\
        \midrule
        \texttt{Blog}+\texttt{Comments}+\texttt{Modu} & 150B & \textit{144.67} & \up{82.82} & \up{54.94} & \up{72.27} & \up{4.09} & \up{21.17} & \up{65.01} \\
        \midrule
        \midrule
        \multicolumn{8}{l}{\textbf{The Case of Adding} \texttt{News} \textbf{into} \texttt{KiN}+\texttt{Ency} \textbf{to Try to Enhance the Performance of YNAT}}\\
        \midrule
        \texttt{News}+\texttt{KiN}+\texttt{Ency} & 101.8B & \textit{142.13} & \up{75.96} & \up{35.42} & \up{55.60} & \up{8.70} & \up{23.38} & \down{27.54} \\
        \bottomrule
    \end{tabular}
    \end{threeparttable}
    \caption{In-context few-shot learning performance with different corpus combination.}
    \label{table:result2}
\end{table*}

\begin{table*}[t!]
    \centering
    \small
    \begin{threeparttable}
    \begin{tabular}{lcccccccc}
        \toprule
        \multirow{2}{*}{Corpus Type} & Corpus & \multirow{2}{*}{PPL} & NSMC & \multicolumn{2}{c}{KorQuAD} & \multicolumn{2}{c}{AI Hub (BLEU)} & YNAT \\
        & Train  & & (Acc)  & (EM) & (F1) & Ko$\rightarrow$En & En$\rightarrow$Ko & (F1) \\
        \midrule
        \midrule
        \texttt{ALL} & 150B & \textit{119.99} & \down{61.88} & \up{56.17} & \up{73.47} & \up{7.43} & \up{24.81} & \up{42.79}  \\
        \midrule
        \texttt{KiN}+\texttt{Ency} & 28.7B & \textit{164.69} & \down{56.78} & \up{42.09} & \up{61.00}  & \up{11.51}  & \up{24.93}  & \up{37.71}  \\
        \texttt{Cafe}+\texttt{KiN} & 109.5B & \textit{141.92} & \down{59.27} & \up{38.45}  & \up{59.00}  & \up{10.12} & \up{24.95} & \up{45.44} \\
        \midrule
        \texttt{Cafe}+\texttt{News} & 150B & \textit{154.20} & \down{66.92} & \down{8.95} & \down{22.85} & \up{3.49} & \up{15.77} & \up{47.34} \\
        \midrule
        \texttt{Blog}+\texttt{Comments}+\texttt{Modu} & 150B & \textit{144.67} & \up{69.15} & \up{54.94} & \up{72.27} & \up{6.06} & \up{22.03} & \up{48.25}  \\
        \midrule
        \texttt{News}+\texttt{KiN}+\texttt{Ency} & 101.8B & \textit{142.13} & \down{61.49} & \up{35.42} & \up{55.60} & \up{10.18}  & \up{24.13}  & \up{51.89}  \\
        \bottomrule
    \end{tabular}
    \end{threeparttable}
    \caption{In-context zero-shot learning performance with different corpus combination.}
    \label{table:result2-zero}
\end{table*}

\begin{table}[t!]
    \centering
    \scriptsize
    \begin{threeparttable}
    \begin{tabular}{lccccc}
        \toprule
         \# of & NSMC & KorQuAD & \multicolumn{2}{c}{AI Hub (BLEU)} & YNAT \\
        tokens & (Acc)  & (EM) & Ko$\rightarrow$En & En$\rightarrow$Ko & (F1) \\
        \midrule
        \midrule
        150B & 84.59 & 56.17 & 6.15 & 23.36 & 59.57  \\
        56B & 84.35 & 55.13 & 5.47 & 22.98 & 51.89 \\
        6B & 74.70 & 36.72 & 3.97 & 17.81 & 30.24 \\
        \bottomrule
    \end{tabular}
    \end{threeparttable}
    \caption{In-context few-shot learning performance of \texttt{ALL} with different size of the pretraining data. The dataset is randomly sampled from the original corpus.}
    \label{table:result3}
\end{table}

\begin{figure}[t]
    \centering
    \includegraphics[width=1.0\linewidth]{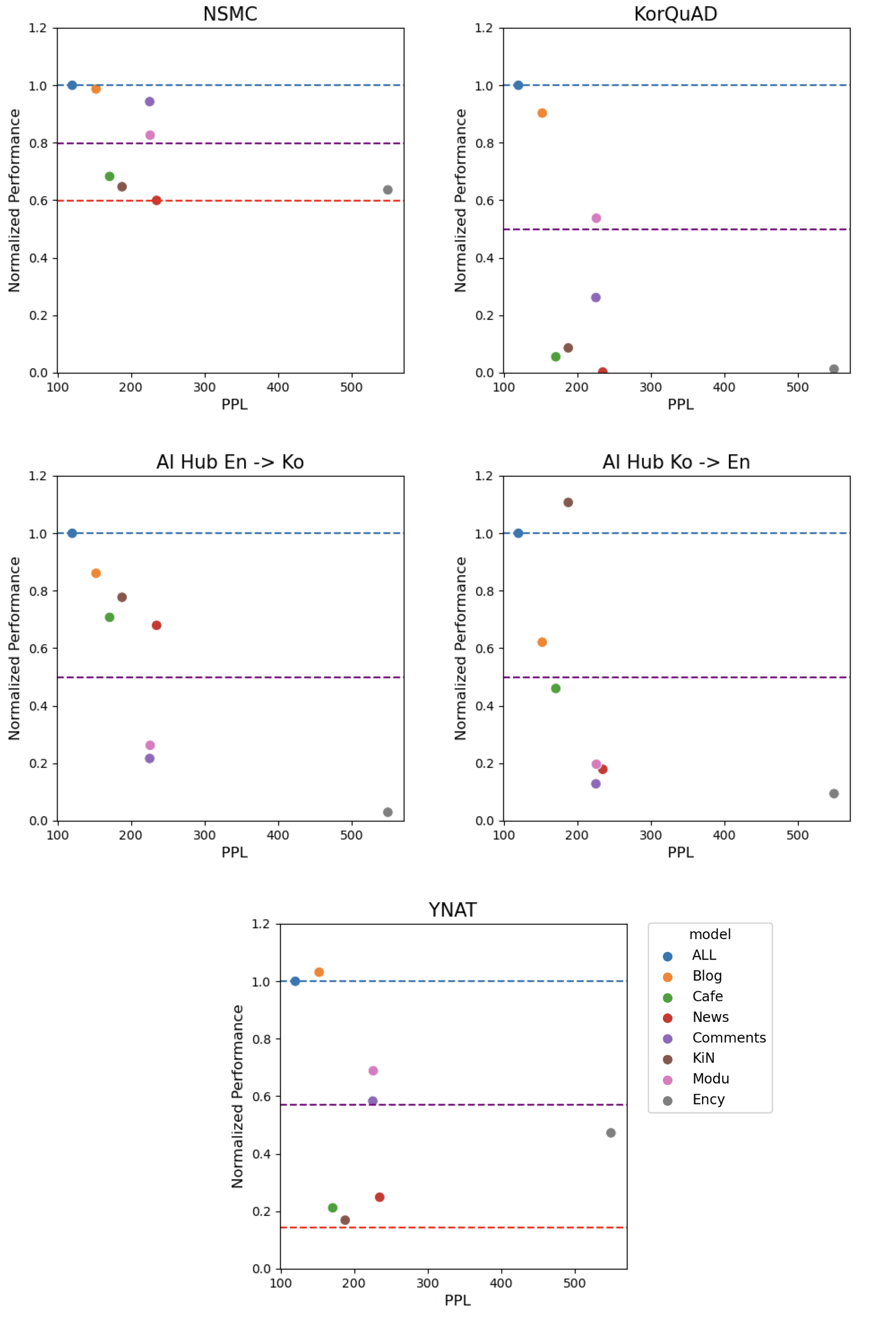}
    \caption{In-context few-shot learning performance of various corpus models and their PPL. A score of the model is divided by that of \texttt{ALL} to calculate the normalized performance. Blue and red lines denote the performance of \texttt{ALL} model and majority baseline, and the purple line represents the average of both defined in the caption of Table~\ref{table:result2}.}
    \label{fig:performance_by_nppl}
\end{figure}

This phenomenon is related to the argument that in-context learning emerges by multi-task learning. According to the argument, as the language modeling objective function requires a language model to learn variety of next word prediction tasks, the generalization pushes in-context learning ability on unseen tasks.
In the example of \texttt{KiN}+\texttt{Ency} model, \texttt{KiN}+\texttt{Ency} may learn in-context learning ability of MRC task, by learning next word prediction tasks of both \texttt{Ency} (Wikipedia) and \texttt{KiN} (QnA).

Unlike these positive cases, we observe that combining corpora does not assure the emergence of competitive in-context learning.
For example, from the case of \texttt{Cafe}+\texttt{News} in Table \ref{table:result2}, even if the mixed corpus model shows slightly better performance on KorQuAD than each of two corpora, its in-context few-shot performances on NSMC, KorQuAD, and YNAT are still below the basis. Furthermore, the performances on NSMC and YNAT even decrease.

\subsection{Effect of Domain Relevance}
\label{subsec:result-relevance}
Speaking of the few-shot results, Table \ref{table:result1} shows that the close relationship between a pretraining corpus and a downstream task does not always guarantee in-context few-shot learning ability on the downstream task. 
\texttt{KiN} and \texttt{Ency} do not perform well on KorQuAD task, although KorQuAD is an MRC task from Korean Wikipedia, \texttt{Ency} includes Korean Wikipedia, and \texttt{KiN} consists of question answering pair, respectively.
Likewise, \texttt{News} does not perform well on YNAT task, although YNAT consists of news headline queries.
Table \ref{table:result2} further shows that \texttt{News}+\texttt{KiN}+\texttt{Ency} model shows more degenerated F1 score on YNAT than \texttt{KiN}+\texttt{Ency}, even though a large amount of \texttt{News} corpus is added to \texttt{News}+\texttt{KiN}+\texttt{Ency} model.

For further investigation, we analyze vocabulary statistics of each corpus. Figure \ref{fig:vocab-overlap} shows the vocabulary overlapping ratio between pretraining corpora and downstream tasks.  
The result shows that high vocabulary overlap between a pretraining corpus and a downstream task does not indicate high downstream task performance. Although the \texttt{Modu} corpus has a large vocabulary overlapping ratio to AI Hub, in-context learning performances of the \texttt{Modu} model on the translation tasks are much lower than \texttt{Blog} and \texttt{KiN}.

The counter example of above supports is AI Hub task performance of \texttt{KiN} model. 
\texttt{KiN} model learned the pattern of Korean-English sentence pairs, since the corpus includes a lot of Korean questions on English language.
While \texttt{KiN} model does not work well in other downstream tasks, the performance on AI Hub translation is competitive and makes the best performance in Ko$\rightarrow$En among seven pretraining corpora. 

In the zero-shot setting, on the other hand, domain relevance seems to affect more positively.
For example, training the \texttt{News} corpus helps in-context zero-shot learning in KLUE-YNAT consistently. As shown in Tables \ref{table:result1-zero} and \ref{table:result2-zero}, the models whose training corpus includes the \texttt{News} corpus (i.e., \texttt{News}, \texttt{Cafe}+\texttt{News}, and \texttt{News}+\texttt{KiN}+\texttt{Ency}) even perform better than the model trained whole HyperCLOVA corpus.

In the case of \texttt{KiN} and AI Hub, zero-shot performance increase for AI Hub tasks of the \texttt{KiN} model is less significant than few-shot. However, adding \texttt{KiN} corpus into the pretraining corpus in the experiments of Table \ref{table:result2-zero} (i.e., \texttt{KiN}+\texttt{Ency}, \texttt{Cafe}+\texttt{KiN}, and \texttt{News}+\texttt{KiN}+\texttt{Ency}) makes a consistent performance increase, and the model outperform \texttt{ALL}.

\subsection{Perplexity and Downstream Task}
\label{subsec:result-perplexity}

\begin{table*}[t!]
    \centering
    \small
    \begin{threeparttable}
    \begin{tabular}{lcccccccc}
        \toprule
                 & Blog & Cafe & News & Comments & KiN & Modu & Ency & All\\
        \midrule
        \texttt{Blog}     & \textit{126.89} & \textbf{201.15} & \textbf{83.98}  & 599.68  & \textbf{138.95} & 98.83  & \textbf{108.17} & \textbf{152.40} \\
        \texttt{Cafe}     & \textbf{168.03} & \textit{135.37} & 107.78 & \textbf{596.27} & 163.94  & 124.71 & 142.15 & 170.85 \\
        \texttt{News}     & 281.33 & 432.90 & \textit{60.21} & 1543.03 & 253.73 & \textbf{87.65}  & 156.28 & 234.78 \\
        \texttt{Comments} & 228.37 & 242.59 & 176.28 & \textit{390.30}  & 164.88 & 196.40 & 239.40 & 225.39 \\
        \texttt{KiN} & 232.13 & 278.89 & 150.78 & 689.89 & \textit{50.06} & 172.78 & 141.45 & 187.80 \\
        \texttt{Modu}     & 267.59 & 411.35 & 84.36  & 1086.04 & 243.19 & \textit{69.48}  & 136.02 & 226.01 \\
        \texttt{Ency} & 841.53 & 1348.38 & 213.87 & 5889.79 & 543.69 & 266.11 & \textit{73.06} & 549.40\\
        \bottomrule
    \end{tabular}
    \end{threeparttable}
    \caption{Validation perplexity scores per each subcorpus. All denotes the validation perplexity on our main validation set from seven corpus sources. \textit{Italic} font denotes the validation PPL of their corpus domain, and \textbf{Bold} denotes second best after own corpus. Overall, \texttt{Blog} has the best overall validation perplexity in most tasks.}
    \label{table:ppl-all}
\end{table*}

\begin{figure}[t]
    \centering
    \includegraphics[width=0.49\linewidth]{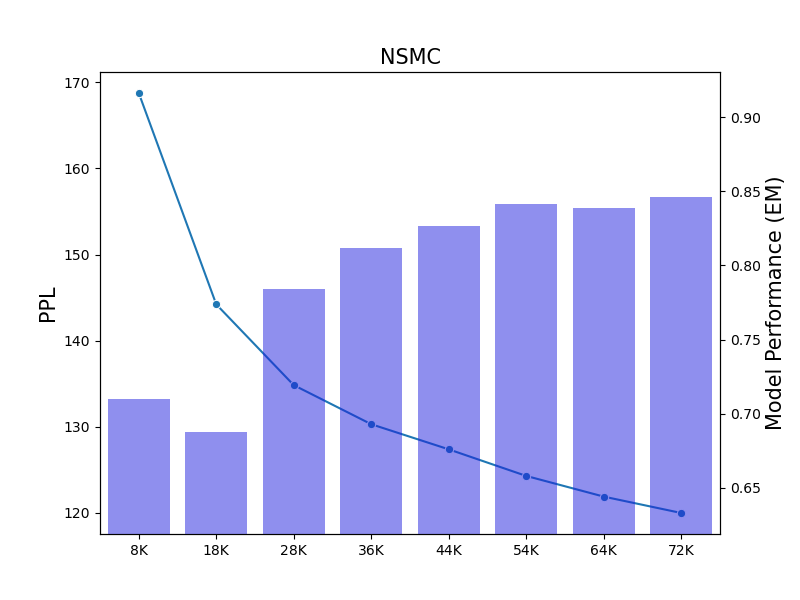}
    \includegraphics[width=0.49\linewidth]{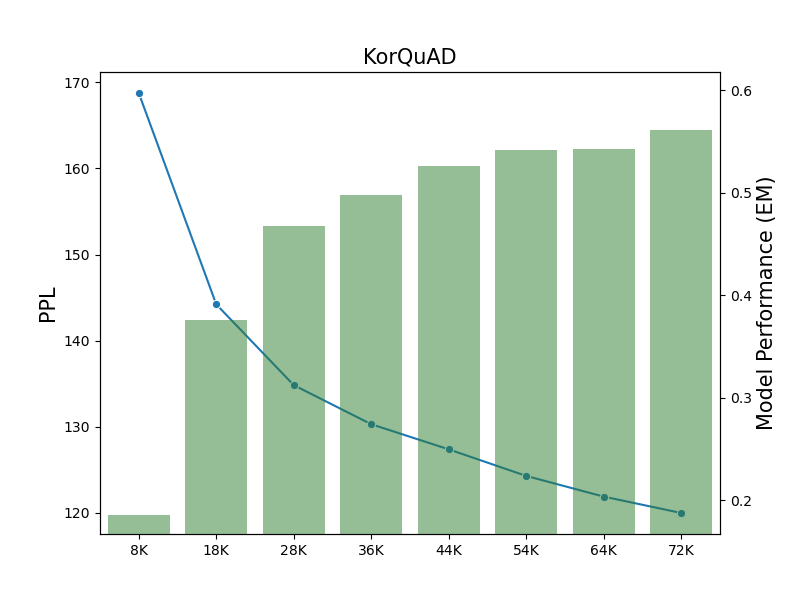}
    \includegraphics[width=0.49\linewidth]{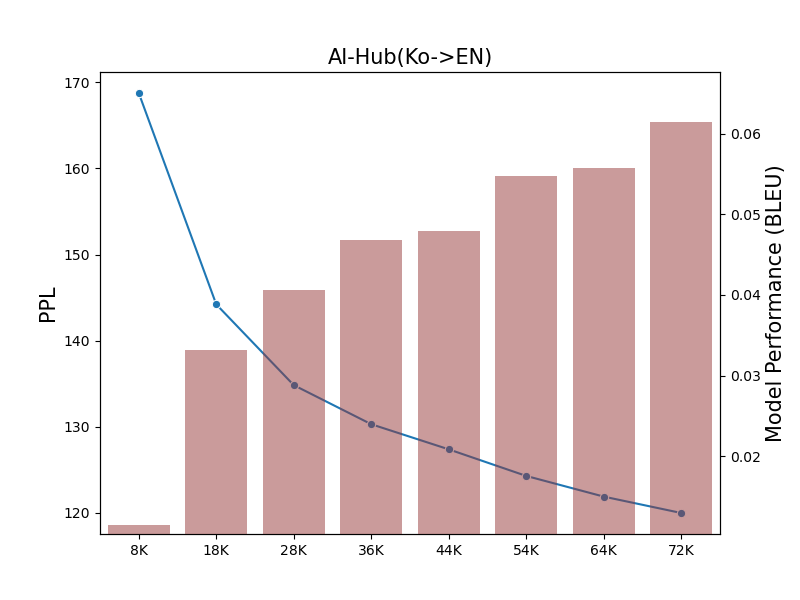}
    \includegraphics[width=0.49\linewidth]{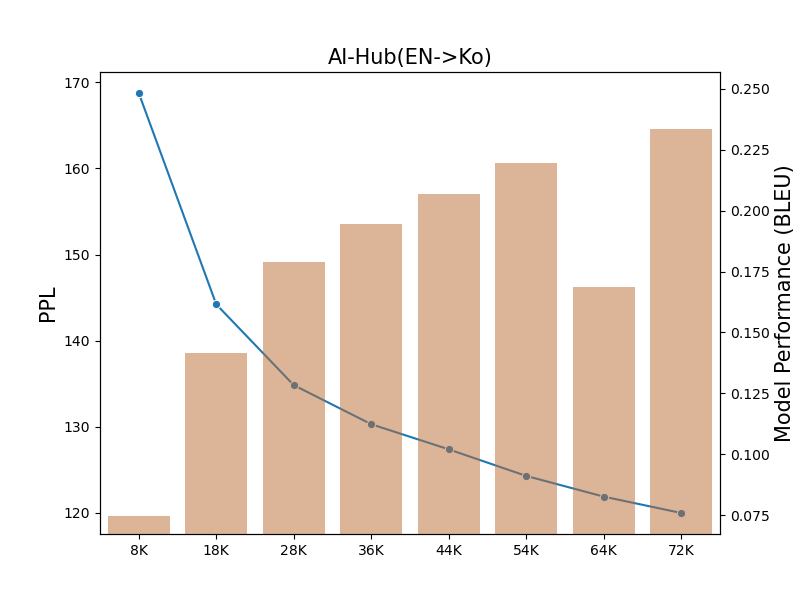}
    \includegraphics[width=0.49\linewidth]{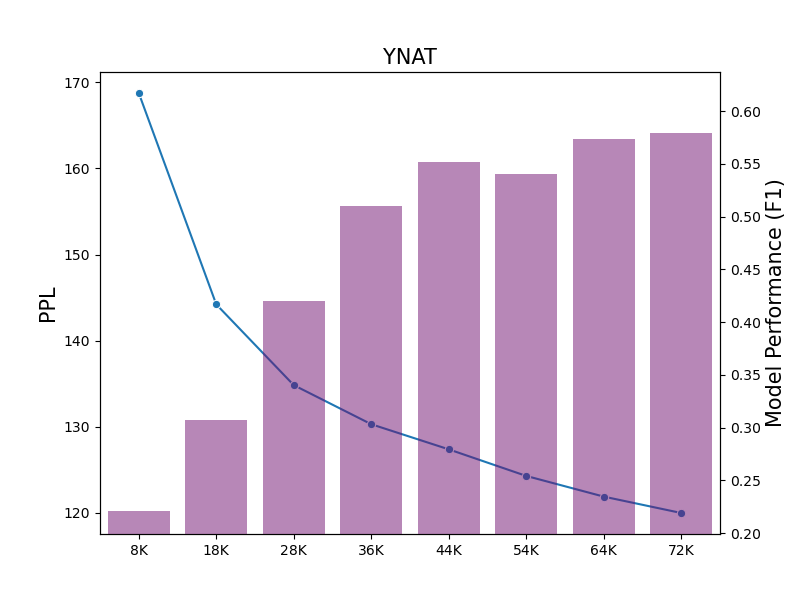}
    \caption{Relation between validation PPL and in-context few-shot learning performance for five downstream tasks as pretraining steps proceed. The results come from the 1.3B-sized \texttt{ALL} model.}
    \label{fig:step_vs_downstream_task}
\end{figure}

Figure~\ref{fig:performance_by_nppl} presents the scatter plots of PPL ($x$-axis) and in-context few-shot learning performance ($y$-axis) on five downstream tasks for single corpus models and the \texttt{ALL} model. 
In Figure \ref{fig:performance_by_nppl}, we normalized in-context few-shot learning performance by dividing \texttt{ALL} model performance for calibrating various task metrics. Because we observe less explicit tendency of correlation between validation perplexity and in-context performance, we argue that it is difficult to hypothesize better perplexity assures emerging of in-context few-shot learning ability.

According to Table \ref{table:result1}, \texttt{Blog} model shows both the lowest PPL and the best in-context learning performance, and \texttt{Ency} model shows both the highest PPL and the worst in-context learning performance.
On the contrary, while \texttt{Cafe} model and \texttt{KiN} model shows the second and third lowest PPL, in-context few-shot learning ability was not observed.
These results show that the perplexity does not serve as a strong predictor of in-context few-shot learning performance in comparing models trained using different corpora.
Table \ref{table:result1} also shows that the corpus size affects in-context few-shot learning performance more than PPL. \texttt{Blog} 27B performs notably worse than \texttt{Blog}, but PPL relatively does not decrease as much.

Speaking of zero-shot results, it seems Table \ref{table:result1-zero} shows that in-context zero-shot learning performances relatively more correlate with perplexity than the few-shot cases. Nevertheless, \texttt{Modu} still has both relatively high perplexity and relatively high in-context zero-shot learning performances.

Table \ref{table:ppl-all} shows validation perplexity scores for each subcorpus. Each row corresponds to the model and each column corresponds to the validation set's subcorpus. Each validation set except All in Table \ref{table:ppl-all} consists of 10,000 instances, and is the part of our main validation sets, consists of 70,000 instances.

On the other hand, Figure \ref{fig:step_vs_downstream_task} shows that PPL and in-context learning performance correlate well in the perspective of training a single model. We can find that the correlation trends between the cases in the training and the cases between the corpus domain are different.

\section{Discussion}

Our knowledge can be used to increase the performance of in-context learning when the corpus is small or/and there exists demand for collecting more corpus.
In the case of XGLM \cite{lin2021few}, which is a concurrent work on multilingual GPT-3, achieved better in-context learning performance for many languages. However, it does not reach the performance of a single language model.
We hope our observation can give insight into what types of pretraining to be collected more, both for multilingual model and low-resource language model.

Another notable example comes from Gopher \cite{rae2021scaling}, which is a concurrent work on state-of-the-art in-context learner.  
\citet{rae2021scaling} determine the ratio between subcorpora based on the perplexity of the validation corpus. They implicitly claim that this ratio results in better downstream task performance, but do not address explicit evidence for this. On the other hand, we are in a position to doubt the strong correlation between perplexity and in-context learning, especially in the few-shot setting.
We hope our findings contribute to making better in-context learners along with other research.

\section{Conclusion}
This paper investigates the effects of the source and the size of the training corpus on in-context learning ability, using the HyperCLOVA corpus. Our discoveries include that corpus sources play a crucial role in whether or not in-context learning ability will emerge in a large-scale language model.

One direction for future work is to investigate linguistic properties of corpus sources which make a competitive in-context learning model. For example, quantifying the difference between two corpora can shed light on how to select suitable corpora for NLP practitioners who build large-scale language models. In addition, intensive studies on different corpus sources other than the HyperCLOVA corpus can help understand the properties of in-context learning.

\section*{Broader Impact Statement}
We present multiple pieces of evidence that models using only a part of the pretraining corpus are comparable with those trained with the entire corpus in terms of in-context performances. Although we leave the validation on larger-scale models, such as tens of billion parameters, to future work, our analysis presents a hint to effectively training LMs with smaller corpora. This approach can contribute to alleviating severe energy consumption issues caused by large-scale LMs. 

Meanwhile, our study relates to the misuse and fairness of large-scale LMs. For example, reweighting domain-specific corpus might cause LMs to be biased inherent in the domain corpus. Therefore, alleviating domain corpus bias would be a valuable future direction.

\section*{Acknowledgment}
The authors thank all the members of CLOVA, AI Lab for devoted supporting and discussion. In particular, they thank Joonsuk Park and Seok Ho Yoon for proofreading.

\bibliography{custom}
\bibliographystyle{acl_natbib}

\clearpage

\appendix

\section{Details on Experimental Results}
\label{sec:appendix-details-experimental-results}

\begin{table*}[t!]
    \centering
    \small
    \begin{threeparttable}
    \begin{tabular}{lcccc}
        \toprule
        \multirow{2}{*}{Model} &  NSMC  & \multicolumn{2}{c}{AI Hub (BLEU)} & YNAT \\
         & (Acc)  & Ko$\rightarrow$En & En$\rightarrow$Ko & (F1) \\
        \midrule
        \midrule
        \texttt{ALL} & 84.59\scriptsize{(1.25)}  & 6.15\scriptsize{(0.16)} & 23.36\scriptsize{(0.33)} & 59.57\scriptsize{(4.30)}  \\
        \texttt{ALL} w/o \texttt{Others} &  84.59\scriptsize{(1.25)}  & 6.14\scriptsize{(0.21)} & 23.21\scriptsize{(0.45)} & {50.76\scriptsize{(11.81)}} \\
        \midrule
        \texttt{Blog}  & \up{83.50}\scriptsize{(2.45)}  & \up{3.82}\scriptsize{(0.10)} & \up{20.11}\scriptsize{(0.79)} & \up{60.68}\scriptsize{(5.75)}  \\
        \texttt{Cafe} & \down{57.77}\scriptsize{(10.28)} &  \down{2.83}\scriptsize{(0.19)} & \up{16.53}\scriptsize{(0.63)} & \down{11.04}\scriptsize{(4.43)} \\
        \texttt{News} &  \down{50.72}\scriptsize{(0.56)} & \down{1.10}\scriptsize{(0.78)} & \up{15.88}\scriptsize{(0.84)} & \down{14.36}\scriptsize{(2.38)} \\
        \texttt{Comments} & \up{79.78}\scriptsize{(2.38)} &  \down{0.79}\scriptsize{(0.01)} & \down{5.06}\scriptsize{(0.16)} & \up{36.17}\scriptsize{(3.31)} \\
        \texttt{KiN} & \down{54.73}\scriptsize{(4.26)}  & \up{6.81}\scriptsize{(0.88)} & \up{18.16}\scriptsize{(0.71)} & \down{9.23}\scriptsize{(1.96)} \\ 
        \texttt{Modu} & \up{69.91}\scriptsize{(8.41)} & \down{1.21}\scriptsize{(0.06)} & \down{6.13}\scriptsize{(0.39)} & \up{43.27}\scriptsize{(6.72)} \\
        \texttt{Ency} & \down{53.81}\scriptsize{(2.22)} &  \down{0.58}\scriptsize{(0.16)} & \down{0.69}\scriptsize{(0.49)} & \down{27.99}\scriptsize{(2.38)} \\
        \midrule
        \texttt{Blog 54B} & \up{83.06}\scriptsize{(2.26)} & \up{3.93}\scriptsize{(0.20)} & \up{21.12}\scriptsize{(0.19)} & \up{57.97}\scriptsize{(5.72)} \\
        \midrule
        \texttt{Blog 27B} & \up{80.27}\scriptsize{(2.28)} &  \up{5.35}\scriptsize{(1.95)} & \up{12.32}\scriptsize{(5.68)} & \up{48.19}\scriptsize{(6.71)} \\
        \texttt{Cafe 27B} & \down{49.91}\scriptsize{(0.29)} &  \up{4.25}\scriptsize{(0.35)} & \up{20.74}\scriptsize{(1.57)} & \down{8.60}\scriptsize{(2.79)} \\
        \texttt{News 27B} & \down{50.64}\scriptsize{(3.26)} & \down{2.42}\scriptsize{(1.41)} & \up{15.78}\scriptsize{(4.21)} & \down{27.20}\scriptsize{(5.86)} \\
        \texttt{Comments 27B} & \up{80.50}\scriptsize{(1.44)} &  \down{1.70}\scriptsize{(0.03)} & \down{3.28}\scriptsize{(0.16)} & \down{25.79}\scriptsize{(7.27)} \\
        \bottomrule
    \end{tabular}
    \end{threeparttable}
    \caption{The results of Table \ref{table:result1} with standard deviation in parentheses.}
    \label{table:result1-std}
\end{table*}

\begin{table*}[t!]
    \centering
    \small
    \begin{threeparttable}
    \begin{tabular}{lcccc}
        \toprule
        \multirow{2}{*}{Corpus Type} & NSMC &  \multicolumn{2}{c}{AI Hub (BLEU)} & YNAT \\
        & (Acc)  & Ko$\rightarrow$En & En$\rightarrow$Ko & (F1) \\
        \midrule
        \midrule
        \texttt{KiN}+\texttt{Ency} &  \down{59.17}\scriptsize{(9.34)} &  \up{8.99}\scriptsize{(0.31)} & \up{23.12}\scriptsize{(0.40)} & \up{42.84}\scriptsize{(9.01)}  \\
        \texttt{Cafe}+\texttt{KiN} & \up{76.42}\scriptsize{(4.68)} &  \up{8.41}\scriptsize{(0.68)} & \up{23.41}\scriptsize{(0.38)} & \up{56.96}\scriptsize{(5.79)}\\
        \midrule
        \texttt{Cafe}+\texttt{News} &  \down{54.15}\scriptsize{(3.90)} &  \up{4.45}\scriptsize{(0.12)} & \up{17.77}\scriptsize{(2.19)} & \down{8.19}\scriptsize{(5.32)} \\
        \midrule
        \texttt{Blog}+\texttt{Comments}+\texttt{Modu} & \up{82.82}\scriptsize{(1.93)} & \up{4.09}\scriptsize{(0.16)} & \up{21.17}\scriptsize{(0.43)} & \up{65.01}\scriptsize{(2.90)} \\
        \midrule
        \texttt{News}+\texttt{KiN}+\texttt{Ency} & \up{75.96}\scriptsize{(5.94)} &  \up{8.70}\scriptsize{(0.45)} & \up{23.38}\scriptsize{(0.18)} & \down{27.54}\scriptsize{(7.46)} \\
        \bottomrule
    \end{tabular}
    \end{threeparttable}
    \caption{The results of Table \ref{table:result2} with standard deviation in parentheses.}
    \label{table:result2-std}
\end{table*}

\begin{table*}[t!]
    \centering
    \small
    \begin{threeparttable}
    \begin{tabular}{lccccccc}
        \toprule
        \multirow{2}{*}{Corpus Type} & Corpus & NSMC & \multicolumn{2}{c}{KorQuAD} & \multicolumn{2}{c}{AI Hub (BLEU)} & YNAT \\
        & Train  & (Acc)  & (EM) & (F1) & Ko$\rightarrow$En & En$\rightarrow$Ko & (F1) \\
        \midrule
        \midrule
        \texttt{ALL} & 150B & 84.59 & 56.17 & 73.47 & 6.15 & 23.36 & 59.57  \\
        \midrule
        \midrule
        \multicolumn{8}{l}{\textbf{The Case where In-context learning Emerges by Combining Two Poor Corpora}} \\
        \midrule
        \texttt{KiN}+\texttt{Ency} & 28.7B & \down{59.17\scriptsize{(-25.42)}} & \up{42.09\scriptsize{(-14.08)}} & \up{61.00\scriptsize{(-12.47)}} & \up{8.99\scriptsize{(+2.84)}} & \up{23.12\scriptsize{(-0.24)}} & \up{42.84\scriptsize{(-16.73)}}  \\
        \texttt{Cafe}+\texttt{KiN} & 109.5B & \up{76.42\scriptsize{(-8.17)}} & \up{38.45\scriptsize{(-17.72)}} & \up{59.00\scriptsize{(-14.47)}} & \up{8.41\scriptsize{(+2.26)}} & \up{23.41\scriptsize{(+0.06)}} & \up{56.96\scriptsize{(-2.61)}}\\
        \midrule
        \midrule
        \multicolumn{8}{l}{\textbf{The Case where In-context learning Does Not Emerge by Combining Two Poor Corpora}} \\
        \midrule
        \texttt{Cafe}+\texttt{News} & 150B &  \down{54.15\scriptsize{(-30.44)}} & \down{22.86\scriptsize{(-33.31)}} & \down{22.72\scriptsize{(-50.75)}} & \up{4.45\scriptsize{(-1.70)}} & \up{17.77\scriptsize{(-5.59)}} & \down{8.19\scriptsize{(-51.38)}} \\
        \midrule
        \midrule
        \multicolumn{8}{l}{\textbf{The Case of Combining In-context Emerging Corpora}}\\
        \midrule
        \texttt{Blog}+\texttt{Comments}+\texttt{Modu} & 150B & \up{82.82\scriptsize{(-1.77)}} & \up{54.94\scriptsize{(-1.23)}} & \up{72.27\scriptsize{(-1.20)}} & \up{4.09\scriptsize{(-2.06)}} & \up{21.17\scriptsize{(-2.19)}} & \up{65.01\scriptsize{(+5.44)}} \\
        \midrule
        \midrule
        \multicolumn{8}{l}{\textbf{The Case of Adding} \texttt{News} \textbf{into} \texttt{KiN}+\texttt{Ency} \textbf{to Try to Enhance the Performance of YNAT}}\\
        \midrule
        \texttt{News}+\texttt{KiN}+\texttt{Ency} & 101.8B & \up{75.96\scriptsize{(-8.63)}} & \up{35.42\scriptsize{(-20.75)}} & \up{55.60\scriptsize{(-17.87)}} & \up{8.70\scriptsize{(+2.55)}} & \up{23.38\scriptsize{(+0.02)}} & \down{27.54\scriptsize{(-32.03)}} \\
        \bottomrule
    \end{tabular}
    \end{threeparttable}
    \caption{Table \ref{table:result2} which includes the difference from \texttt{ALL} in parentheses.}
    \label{table:result2-difference}
\end{table*}

\begin{figure}[t]
    \centering
    \includegraphics[width=1.0\linewidth]{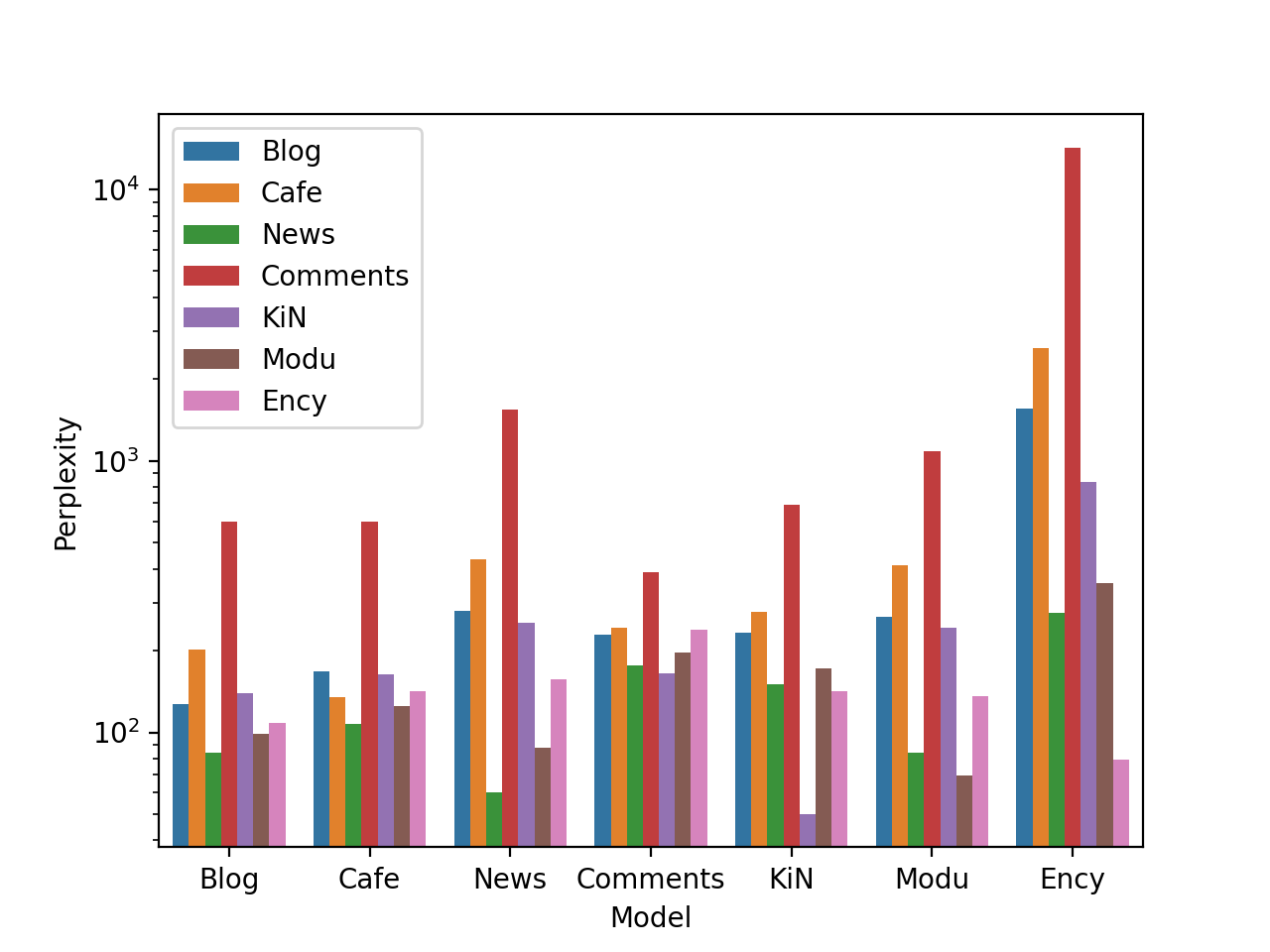}
    \caption{Validation perplexity of different 1.3B-size models in log scale. Color indicates the source of  validation corpus.}
    \label{fig:ppl}
\end{figure}

Tables \ref{table:result1-std} and \ref{table:result2-std} show standard derivation value on Tables \ref{table:result1} and \ref{table:result2}.
Table \ref{table:result2-difference} shows score difference with \texttt{ALL} in addition to in-context learning scores on Table \ref{table:result1}.
Figure \ref{fig:ppl}, supporting Table \ref{table:ppl-all}, shows the validation perplexity of different model from different corpus.

\section{Details on Pretraining Corpus}

Tables \ref{table:corpus-example} and \ref{table:corpus-example-en} show example
instances of seven pretraining corpus in Korean and English, respectively.

For preprocessing steps of our pretraining corpus, we use HyperCLOVA corpus which is also used in \cite{kim2021changes} as described in Section 3.1. Therefore, we share the preprocessing steps of \citet{kim2021changes}. Appendix A in \cite{kim2021changes} describes their preprocessing methods on data descriptoin, data clearning, data anonymization, and data postprocessing.

\subsection{Deduplication Preprocess}
\label{subsec:dedup}
We additionally introduce the deduplication preprocess of HyperCLOVA corpus, which is used in \cite{kim2021changes}. 
The deduplication preprocess was applied to construct HyperCLOVA corpus to prevent explicit duplication within and between subcorpora \cite{kim2021changes}. According to the response of \citet{kim2021changes}, they use an in-house search engine and an in-house engineering trick to detect document pairs that are very similar to each other. There are two pipelined steps: (1) removing duplicates within subparts of the corpus, and then (2) removing duplicates between subparts of the corpus. Therefore, documents with high overlap do not exist throughout the documents.
Here, the number of subparts is 29. These 29 subparts are categorized into the eight domains we deal with in the paper (i.e., \texttt{Blog}, \texttt{News}, \texttt{Cafe}, \texttt{Comments}, \texttt{KiN}, \texttt{Modu}, \texttt{Ency}, and \texttt{Others}).
Overall, there is no explicit overlap between each corpus, since very similar documents have already been removed from the corpus.
The overlap between eight HyperCLOVA subcorpora is quite small. There were many overlaps within the subpart of the corpus. However, the overlap between subparts of the corpus was only 0.024\% of the total, according to the counts in the second pipelined step of deduplication between subparts.

\section{Experiments on LoRA}
\label{sec:lora}

\begin{table}[t!]
\centering
\begin{threeparttable}
\begin{tabular}{lcc}
\toprule
\multirow{2}{*}{Model} & NSMC & YNAT \\
        & (Acc)   & (F1)         \\
\midrule
\midrule
\texttt{ALL} & 91.83 & 86.47 \\
\midrule
\texttt{Comments} & 92.02  & 84.07     \\
\texttt{Blog}     & 91.93  & 86.21      \\
\texttt{Cafe}     & 91.57     & 85.45      \\
\texttt{News}     & 90.62      & 86.57      \\
\texttt{KiN}      & 90.89    & 84.46   \\
\texttt{Modu}     & 90.60   & 86.31 \\
\texttt{Ency}     & 86.93     & 82.37   \\
\midrule
\texttt{KiN}+\texttt{Ency} & 90.92  & 84.00  \\   
\texttt{Cafe}+\texttt{KiN} & 90.99  & 87.52     \\  
\texttt{Cafe}+\texttt{News}        & 91.37     & 86.37   \\ 
\texttt{Blog}+\texttt{Comment}+\texttt{Modu}  & 88.83  & 87.07 \\
\texttt{News}+\texttt{KiN}+\texttt{Ency}  & 91.13     & 86.62 \\ 
\bottomrule
\end{tabular}
\end{threeparttable}
\caption{LoRA finetuning performance on different pretraining corpus and its combination.}
\label{table:lora}
\end{table}

Table \ref{table:lora} shows the results of LoRA \cite{hu2021lora} finetuning on some models in Tables \ref{table:result1} and \ref{table:result2}.

\section{Examples of Few-shot Prompt}
\label{sec:appendix-examples-of-few-shot-prompt}
Tables \ref{table:nsmc-few-shot-example-ko}, \ref{table:korquad-few-shot-example-ko}, \ref{table:translation-few-shot-example-ko}, and \ref{table:ynat-few-shot-example-ko} show the example few-shot prompt of NSMC, KorQuAD, AI Hub, and YNAT, respectively.
Tables \ref{table:nsmc-few-shot-example-en}, \ref{table:korquad-few-shot-example-en}, and \ref{table:ynat-few-shot-example-en} show the translated version for NSMC, KorQuAD, and YNAT, respectively.

On the other hand, the number of random seed is one for KorQuAD. We explain why evaluation on KorQuAD with many random seeds is difficult, from the perspective of prompt design. The way we make randomness on trials is to change few-shot examples in the prompt. However, in the case of \citet{kim2021changes} and in our case, there are no alternative examples to put into the prompt. The prompt examples of KorQuAD are one document and a few question-answer pairs, and not a few document-question-answer triples.
In other words, in the prompt of KorQuAD, the number of the document is one. Thus, the document is used for both few-shot question-answer pairs and a query question for the inference. In KorQuAD, there are five corresponding question-answer pairs in each document. In the experimental setting of ours and \citet{kim2021changes}, four question-answers are put into the prompt and one question is used for the test. Therefore, there are no other question-answer pairs to replace the four pairs.

\section{Generalization to Other Languages}
Someone can ask whether our results can be extended to other languages, including English. We have left experiments on non-Korean language as future work.
However, we describe some explanations below to defend our experiments on the Korean language and to discuss why experiments on other languages are practically non-trivial.

First, we think our findings are basically generalizable to other languages. From the perspective of pretraining and in-context learning, fundamental differences between Korean and English were limitedly reported.
For example, XGLM \cite{lin2021few}, a concurrent work on, also does not show critical evidence on language-specific properties.

Second, It is non-trivial to control various aspects of corpora for our purpose. Most corpus for in-context few-shot learners comes from crawled website which is not easy to distinguish from its original source. For example, 82\% of OpenAI GPT-3 Corpus \cite{brown2020language} is a filtered version of Common Crawl. In this regard, we used relatively a well-refined corpus which consist of several subcorpus from a single web service. (Please see also Section \ref{subsec:dedup} of this letter.) On the other hand, we have interests to extend our work onto Pile dataset~\cite{gao2020pile}, by controlling the subcorpora in the direction our study pursuits, in the future.

\begin{table*}[t!]
    \centering
    \small
    \begin{tabular}{lp{135mm}}
        \toprule
                 & Example\\
        \midrule
        \texttt{Blog}
        & 블로그 \newline제목: 촬영하러 온 꼬맹이들\^{}\^{} \newline본문: 엄마 회사에 오늘은 모델로 일하러온 꼬맹이들. 신나게 놀고 까불고 뛰어다니다가 책보더니 잠잠해진다. 조용해진 아이들 보고 놀라는 스탭들. 순간 엄마얼굴엔 미소가! ㅋㅋ. 책 잘 읽는 아이들이라 자랑스러움이 잠시 ㅋㅋ 그러다가 촬영하고 생각보다 잘해줘서 고맙네. 엉망진창으로 못할줄 알았는데 카메라를 보다니. ㅋ 어린시절부터 카메라를 본 경험이 빛을 발하긴 하나보다. ㅋㅋ 하여간. 엄마 회사에 모델로 와준 꼬맹이들. 고마워\^{}\^{}. 좋은 추억이 되었길.\\
        \midrule
        \texttt{Cafe}     & 카페 \newline 제목: 탐스우먼 상자채 새신발(직구한것보다 싸게 내놓아요~~ \^{}\^{}) \newline본문: 벼룩시장(중고), 판매중, 가격 1원, 안전거래 미사용, 탐스클래식, 판 매 양 식아이디 이메일싸이,블로그,타카페,타사이트 링크시 삭제 및 강퇴 거주지역도,시,동까지 정확히 기재 판매 제품명 구입시기년,월 기재 희망가격정확히 기재: (3만~4만등의 경매 유도글 삭제) 거래방법 직거래, 택배, 안전거래 상세설명 탐스 공홈에서 직구했는데사이즈가 커서 내놓아요~~ 빨리 팔려구 직구한것 보다 싸게 내놓아요~~\^{}\^{} 1. 탐스우먼 유니버시티 애쉬 그레이 택포 45,00 2. 탐스우먼 초코캔버스 택포 40,000 많은 문의 부탁드려요~~\^{}\^{} \\
        \midrule
        \texttt{News}     & 뉴스 \newline제목: 전명환, 이병기 시문학상 수상 \newline본문: `2016 이병기 청년시문학상ㆍ최명희 청년소설문학상' 수상자가 결정됐다. 지난 1일 전북대 총장실에서 시상식을 연 가운데 이병기 청년시문학상 대학 부문에는 `대과거'를 쓴 전명환(중앙대 국어국문 2년), 고등 부문에는 `몽상'을 선보인 황주연(경산여고 2년) 이 선정됐다. 최명희 청년소설문학상 대학 부문에는 `꽃에서부터'를 쓴 윤선미(서울디지털대 문창 3년), 고등 부문에는 `야간비행'을 쓴 윤정은(안양예고 2년)이 수상의 영예를 안았다. 전북대학교(총장 이남호) 신문방송사와 혼불기념사업회ㆍ최명희문학관(대표 장성수)이 공동으로 주관하는 공모전에는 올해 시 부문 167명 669편, 소설 부문 108명 116편이 출품됐다. 시 부문 심사는 최승범 양병호 유인 이승철 위원이, 소설 부문 심사는 서철원 황보윤 장마리 김소윤 최기우 위원이 맡았다. 박준호 문학상 운영위원장 및 신문방송사 주간은 "수준 높았으며 시대를 바라보는 청년들의 녹록치 않은 고민과 생각을 엿볼 수 있었다"고 평했다. \\
        \midrule
        \texttt{Comments} & 대화 \newline본문: 하루를 엄청 길게 사용하시네요\^{}\^{} 점심은 더 많이 드세요~~ \newline아점입니다ㅎㅎ 저녁을 기다려야죠\^{}\^{} \newline이렇게 드시고 무슨 운동까지 하십니까?? 저녁 윗몸일으키기는 빼세요~~ \newlineㅋㅋㅋ요즘 가끔 빼먹습니다..저 담주 월.화중에 앤더슨님 방문할까합니다..같이가시죠? \newline다음주 월,화요??\^{}\^{} 가야죠,,갑니다..시간을 만들어서라도 가야죠\^{}\^{} 어찌 움직이실건지요?? \newline화요일날로..저는 전철을타야해서 무찌르자님은 어디서 출발하시는지요 \newline전 서울 성수동에서 출발합니다.. 성수까지만 오시면 제가 모시겠습니다..\^{}\^{} \\
        \midrule
        \texttt{KiN} & 질의응답 \newline질문: 독실라게???? 사투리라는데 독실라게 가 뭔뜻인가요? 경상도 쪽이라는데. \newline본문: 경상도 방언에서는 엄청나게 .억수로 강조하는 부사입니다\\
        \midrule
        \texttt{Modu}     & 뉴스 \newline본문: 춘분에 눈 내린 부산...강풍까지 불며 피해 속출눈이 잘 오지 않는 부산에 춘분인 21일 0.7cm의 눈이 내려 산간지역 도로가 통제되는 등 피해가 잇따랐다고 연합뉴스가 보도했다. 부산기상청에 따르면 이날 부산의 아침 최저기온은 공식 관측소가 있는 중구 대청동 기준 1도였다. 해발고도가 500m 이상인 중구 서대신동은 영하 1.4도, 영도구는 영하 0.9도를 기록했다. 강한 바람까지 불면서 체감온도는 영하 2.6도까지 떨어졌다. 아침 최저기온이 영하권을 넘나들면서 밤새 내리던 비가 진눈깨비로 변했다. 부산 기장군에 있는 기상청 적설 자동 관측장비에는 적설량이 0.7cm로 기록됐다.\\
        \midrule
        \texttt{Ency} & 문서 \newline 제목: 설악면 \newline 본문: 설악면(雪岳面)은 대한민국 경기도 가평군의 면이다. 넓이는 141.53 km²이고, 인구는 2016년 12월 말 주민등록 기준으로 8,986 명이다.설악면은 북한강 남쪽에 있어서 본래 양평군에 속했는데, 1942년 가평군에 편입되었다. 강원도에 있는 설악산(雪嶽山)과는 무관하다. \\
        \bottomrule
    \end{tabular}
    \caption{Example document from various domains. Note that \texttt{Modu} consists of 5 different subdomains and the example is taken from the news subdomain, which is the largest.}
    \label{table:corpus-example}
\end{table*}

\begin{table*}[t!]
    \centering
    \small
    \begin{tabular}{lp{135mm}}
        \toprule
                 & Example\\
        \midrule
        \texttt{Blog}
        & Blog \newline Title: Kids who came to shoot \^{}\^{} \newline Body: Kids who came to my mom's company as models today. After having fun, playing around, and running around, they read a book and calmed down. Staff members are surprised to see the quiet kids. A smile on my mom's face! Haha. I'm proud of them because they're good at reading books, and then I took phots and thank them for doing better than I thought. I thought they do mess it up, but I can't believe that they are looking at the camera. The experience of looking at the camera since early childhood must have helped. Anyway. The kids who came to my mom's company as models. Thank you \^{}\^{}. I hope it was a good memory. \\
        \midrule
        \texttt{Cafe}     & Cafe \newline Title: Toms women's shoes (It's cheaper than what I bought directly~\^{}\^{}) \newline Text: Flea market (used), selling, price of 1 won, not used for safety transactions, Tom's Classic, sales form ID Email, blog, other cafe, other site link, city, dong, exact date of purchase of sales product, desired price of 30,000 to 40,000 won, direct auction transaction. Urgent sale and sell it cheaper than what I bought directly.~\^{}\^{}1. Tom's Woman University Ash Gray 45,00 2. Tom's Woman Chocolate Canvas 40,000. Please contact us.\^{}\^{} \\
        \midrule
        \texttt{News}     & News \newline Title: Jeon Myeonghwan and Lee Byungki won the Poem Literature Award. \newline Body: The winners of the 2016 Young Poetry Literature Award and Choi Myung-hee Young Novel Literature Award have been decided. While the awards ceremony was held at Jeonbuk National University's president's office on the 1st, Jeon Myung-hwan (second year of Chung-Ang University's Korean Language Language) who wrote "the past" in the college category and Hwang Joo-yeon (second year of Gyeongsan Girls' High School) were selected. Yoon Sun-mi (3rd year of Moonchang, Seoul Digital University), who wrote "From Flowers" in the college category of Choi Myung-hee's Youth Novel Literature Award, and Yoon Jung-eun (2nd year of Anyang Arts High School), who wrote "Night Flight" in the high school category, were honored. The contest, co-hosted by Jeonbuk National University (President Lee Nam-ho) newspaper broadcasters, Honbul Memorial Society, and Choi Myung-hee Literature Museum (CEO Jang Sung-soo), featured 669 works of 167 people in the poetry category and 116 works of 108 people in the novel category this year. Choi Seung-beom, Yang Byung-ho, Yoo Seung-chul, a member of the Yoo In, and Seo Cheol-won, Hwang Bo-yoon, Jang Mari, Kim So-yoon, and Choi Ki-woo, a member of the novel division, were in charge of the screening. Park Joon-ho, chairman of the Literature Award's steering committee and weekly newspaper broadcaster, commented, "It was high-quality, and I could get a glimpse of the difficult worries and thoughts of young people looking at the times." \\
        \midrule
        \texttt{Comments} & Conversation \newline Body: You spend a long day.\^{}\^{} Eat more for lunch.~~ \newline It's brunch. We have to wait for dinner.\^{}\^{} \newline What kind of exercise do you do after eating like this? Don't do sit-ups in the evening.~~ \newline I've been skipping it from time to time. I'm going to visit Anderson next Monday and Tuesday.Let's go together, right? \newline Next Monday and Tuesday?\^{}\^{} I have to go, I'm going...I'll make time to go there.\^{}\^{} How are you going to move? \newline On Tuesday... I have to take the subway, so where will you leave? \newline I'm departing from Seongsu-dong, Seoul. If you come all the way to Seongsu, I'll take you.\^{}\^{} \\
        \midrule
        \texttt{KiN} & QnA \newline Question: Doksilagae? It's a dialect. What does doksilagae mean? It's used near Gyeongsang-do. \newline Text: In Gyeongsang-do dialect, it means tremendously.It's an adverb to emphasize.\\
        \midrule
        \texttt{Modu}     & News \newline Text: It snowed in the spring equinox in Busan...Strong winds are blowing and they're avoiding it. Yonhap News Agency reported that 0.7 centimeters of snow fell on the 21st, the spring equinox in Busan, where snow was not easy, and roads in mountainous areas were controlled. According to the Busan Meteorological Administration, the lowest temperature in the morning in Busan was 1 degree in Daecheong-dong, Jung-gu, where the official observation station is located. Seodaemun-dong, Jung-gu, with an altitude of more than 500m above sea level, recorded minus 1.4 degrees Celsius and Yeongdo-gu recorded minus 0.9 degrees Celsius. As strong winds blew, the sensible temperature dropped to minus 2.6 degrees Celsius. As the morning low temperature crossed below zero, the rain that had been falling all night turned into sleet. 
        The automatic snow observation equipment of the Korea Meteorological Administration in Gijang-gun, Busan recorded a snowfall of 0.7cm. \\ 
        \midrule
        \texttt{Ency} & Document\newline Title: Seorakmyeon. \newline Body: Seorak-myeon is a myeon of Gapyeong-gun, Gyeonggi-do, Korea. The area is 141.53 km², and the population is 8,986 based on resident registration at the end of December 2016. Seorak-myeon was originally part of Yangpyeong-gun in the south of the Bukhangang River, but was incorporated into Gapyeong-gun in 1942. It has nothing to do with Seoraksan Mountain in Gangwon-do. \\
        \bottomrule
    \end{tabular}
    \caption{An example document in Table \ref{table:corpus-example}, translated into English by a machine translator.}
    \label{table:corpus-example-en}
\end{table*}

\begin{table*}[t!]
    \centering
    \small
    \begin{tabular}{lp{115mm}}
        \toprule
        \texttt{Context ->} & 아 더빙.. 진짜 짜증나네요 목소리 (부정) \newline
        흠...포스터보고 초딩영화줄....오버연기조차 가볍지 않구나 (부정) \newline
        너무재밓었다그래서보는것을추천한다 (긍정) \newline
        교도소 이야기구먼 ..솔직히 재미는 없다..평점 조정 (부정) \newline
        사이몬페그의 익살스런 연기가 돋보였던 영화!스파이더맨에서 늙어보이기만 했던 커스틴 던스트가 너무나도 이뻐보였다 (긍정) \newline
        ...\newline
        원작의 긴장감을 제대로 살려내지못했다. \newline \\
        \midrule
        \texttt{Correct Answer ->} & (부정) \\
        \midrule
        \texttt{Incorrect Answer ->} & (긍정) \\
        \bottomrule
    \end{tabular}
    \caption{Formatted dataset example for NSMC. (few-shot: 70)}
    \label{table:nsmc-few-shot-example-ko}
\end{table*}

\begin{table*}[t!]
    \centering
    \small
    \begin{tabular}{lp{115mm}}
        \toprule
        \texttt{Context ->} & Ah dubbing.. It's really annoying. voice (Negative)\newline
        Hm... I saw the poster and gave elementary school student movie lines...Even overacting isn't light. (negative)\newline
        It was so much fun, so I recommend watching it (negative)\newline
        It's about the prison...Honestly, it's not fun.Adjusting the rating (negative)\newline
        It's a movie where Simon Peg's humorous acting stood out!Kirsten Dunst, who only looked old in Spider-Man, looked so pretty. (positive)\newline
        ...\newline
        It did not capture the tension of the original work properly. \newline \\
        \midrule
        \texttt{Correct Answer ->} & (negative) \\
        \midrule
        \texttt{Incorrect Answer ->} & (positive) \\
        \bottomrule
    \end{tabular}
    \caption{An example document in Table \ref{table:nsmc-few-shot-example-ko}, translated into English by a machine translator}
    \label{table:nsmc-few-shot-example-en}
\end{table*}

\begin{table*}[t!]
    \centering
    \small
    \begin{tabular}{lp{110mm}}
        \toprule
        \texttt{Context ->} & 제목: 임종석\newline
        지문: 1989년 2월 15일 여의도 농민 폭력 시위를 주도한 혐의(폭력행위등처벌에관한법률위반)으로 지명수배되었다.
        1989년 3월 12일 서울지방검찰청 공안부는 임종석의 사전구속영장을 발부받았다.
        같은 해 6월 30일 평양축전에 임수경을 대표로 파견하여 국가보안법위반 혐의가 추가되었다.
        경찰은 12월 18일~20일 사이 서울 경희대학교에서 임종석이 성명 발표를 추진하고 있다는 첩보를 입수했고, 12월 18일 오전 7시 40분 경 가스총과 전자봉으로 무장한 특공조 및 대공과 직원 12명 등 22명의 사복 경찰을 승용차 8대에 나누어 경희대학교에 투입했다.
        1989년 12월 18일 오전 8시 15분 경 서울청량리경찰서는 호위 학생 5명과 함께 경희대학교 학생회관 건물 계단을 내려오는 임종석을 발견, 검거해 구속을 집행했다.
        임종석은 청량리경찰서에서 약 1시간 동안 조사를 받은 뒤 오전 9시 50분 경 서울 장안동의 서울지방경찰청 공안분실로 인계되었다.\newline
        질문: 1989년 6월 30일 평양축전에 대표로 파견 된 인물은?\newline
        답변: 임수경\newline
        질문: 임종석이 여의도 농민 폭력 시위를 주도한 혐의로 지명수배된 연도는?\newline
        답변: 1989년\newline
        질문: 임종석을 검거한 장소는 경희대 내 어디인가?\newline
        답변: 학생회관 건물 계단\newline
        질문: 임종석이 조사를 받은 뒤 인계된 곳은 어딘가?\newline
        답변: 서울지방경찰청 공안분실\newline
        질문: 1989년 2월 15일 여의도 농민 폭력 시위를 주도한 혐의로 지명수배된 사람의 이름은?\newline
        답변: \\
        \midrule
        \texttt{Target Completion ->} & 임종석 \\
        \bottomrule
    \end{tabular}
    \caption{Formatted dataset example for KorQuAD: Machine Reading Comprehension (MRC) (few-shot: 4)}
    \label{table:korquad-few-shot-example-ko}
\end{table*}

\begin{table*}[t!]
    \centering
    \small
    \begin{tabular}{lp{110mm}}
        \toprule
        \texttt{Context ->} & Title: Lim Jongseok.\newline
        Main Text: On February 15, 1989, he was wanted for leading a violent demonstration against farmers in Yeouido (violence of the Punishment of Violence, etc. Act).
        On March 12, 1989, the Ministry of Public Security of the Seoul District Prosecutors' Office received a preliminary arrest warrant for Lim Jong-seok.
        On June 30 of the same year, Lim Soo-kyung was dispatched as a representative to the Pyongyang Festival, adding charges of violating the National Security Act.
        The police obtained information that Lim Jong-seok was pushing for a statement at Kyung Hee University in Seoul between December 18 and December 18, and distributed 22 plainclothes police, including 12 special forces and anti-aircraft staff armed with gas guns and electronic rods, to Kyung Hee University.
        At around 8:15 a.m. on December 18, 1989, the Seoul Cheongnyangni Police Station found Lim Jong-seok, who came down the stairs of the Kyunghee University Student Center building with five escort students, arrested him and executed his arrest.
        Lim Jong-seok was investigated by the Cheongnyangni Police Station for about an hour and handed over to the Seoul Metropolitan Police Agency's public security loss office in Jangan-dong, Seoul, at around 9:50 a.m.\newline\newline
        Question: Who was dispatched as a representative at the Pyongyang Festival on June 30, 1989?\newline
        Answer : Lim Su-kyung\newline
        Question: When was Lim Jong-seok wanted to be arrested for leading a violent demonstration against farmers in Yeouido?\newline
        Answer: 1989\newline
        Question: Where in Kyung Hee University did you arrest Lim Jong-seok?\newline
        Answer: Stairs in the building of the student center.\newline
        Question: Where was Lim Jongseok handed over after being investigated?\newline
        Answer: Seoul Metropolitan Police Agency lost public security.\newline
        Question: What is the name of the person who was wanted for leading the Yeouido peasant violence protest on February 15, 1989?\newline
        Answer: \\
        \midrule
        \texttt{Target Completion ->} & Lim Jongseok. \\
        \bottomrule
    \end{tabular}
    \caption{Example document in Table \ref{table:korquad-few-shot-example-ko}, translated into English by a machine translator}
    \label{table:korquad-few-shot-example-en}
\end{table*}

\begin{table*}[t!]
    \centering
    \small
    \begin{tabular}{lp{110mm}}
        \toprule
        \texttt{Context ->} & 스키너가 말한 보상은 대부분 눈으로 볼 수 있는 현물이다.=Skinner's reward is mostly eye-watering.\newline
        심지어 어떤 문제가 발생할 건지도 어느 정도 예측이 가능하다.=Even some problems can be predicted.\newline
        ...\newline
        오직 하나님만이 그 이유를 제대로 알 수 있을 겁니다.=\\
        \midrule
        \texttt{Target Completion ->} & Only God will exactly know why. \\
        \bottomrule
    \end{tabular}
    \caption{Formatted dataset example for AI-Hub: Translation}
    \label{table:translation-few-shot-example-ko}
\end{table*}

\begin{table*}[t!]
    \centering
    \small
    \begin{tabular}{lp{115mm}}
        \toprule
        \texttt{Context ->} & 네이버랩스 3D 지도 기술업체 에피폴라 인수(과학)\newline
        野 北 축구 생중계 거부에 대북정책 현주소종합(정치)\newline
        즐라탄 행선지 정한 듯…큰 소식 알려드리겠다(스포츠)\newline
        페루 아마존 지역서 70대 英 환경운동가 불에 타 숨진 채 발견(세계)\newline
        머리 맞댄 경제부총리와 한국은행 총재(경제)\newline
        전주 MBC 전북 출>신 故 이용마 기자 추모공간 사흘간 운영(사회)\newline
        ...\newline
        구글 인공지능 다음 도전은 스타크래프트\\
        \midrule
        \texttt{Correct Answer ->} & (과학) \\
        \midrule
        \texttt{Incorrect Answer ->} & (세계) \\
        \bottomrule
    \end{tabular}
    \caption{Formatted dataset example for YNAT: Topic Classification (few-shot: 70)}
    \label{table:ynat-few-shot-example-ko}
\end{table*}

\begin{table*}[t!]
    \centering
    \small
    \begin{tabular}{lp{115mm}}
        \toprule
        \texttt{Context ->} & NAVER LABS acquires 3D map technology company Epipolar(Science)\newline
        North Korea's refusal to broadcast live soccer matches the current state of North Korea policy(Politics)\newline
        It seems that Zlatan's destination has been decided... I'll tell you the big news(Sport)\newline
        British environmentalist 70-year-old found burnt to death in Peruvian Amazon(World)\newline
        Deputy Prime Minister of Economy and Bank of Korea Governor(Economy)\newline
        Jeonju MBC Operates a three-day memorial space for the late reporter Lee Yong-ma from Jeonbuk(Social)\newline
        ...\newline
        Google's next AI challenge is Starcraft\\
        \midrule
        \texttt{Correct Answer ->} & (Science) \\
        \midrule
        \texttt{Incorrect Answer ->} & (World) \\
        \bottomrule
    \end{tabular}
    \caption{Example document in Table \ref{table:ynat-few-shot-example-ko}, translated into English by a machine translator}
    \label{table:ynat-few-shot-example-en}
\end{table*}

\end{document}